\documentclass[10pt,journal,compsoc]{IEEEtran}
\usepackage{amsmath,amsfonts}
\usepackage{algorithmic}
\usepackage{algorithm}

\usepackage{array}
\usepackage[caption=false,font=normalsize,labelfont=sf,textfont=sf]{subfig}
\usepackage{textcomp}
\usepackage{stfloats}
\usepackage{url}
\usepackage{verbatim}
\usepackage{graphicx}
\usepackage{cite}
\usepackage{color}
\usepackage{multirow} %
\usepackage{bigstrut} %
\usepackage{nicematrix} %
\usepackage{colortbl} %
\usepackage{bbding}

\usepackage{marvosym} %
\usepackage{booktabs}
\usepackage{hyperref}
\hypersetup{
colorlinks=true, %
linkcolor=black,
anchorcolor=black,
citecolor=black
}
\usepackage{soul} %
\soulregister\cite7 %
\soulregister\ref7 
\usepackage{pifont} %

\definecolor{myGreen}{rgb}{0.78,0.92,0.73}
\definecolor{goodGreen}{rgb}{0,  0.69,  0.314}

\definecolor{yellow}{rgb}{0.99,0.99,0.70}
\definecolor{white}{rgb}{1.0,1.0,1.0}
\definecolor{black}{rgb}{0.00,0.00,0.00}

\newcommand{\para}[1]
{\vspace{.05in}\noindent\textbf{#1}}

\newcommand{\yq}[1]{{\color[rgb]{0.0,0.0,0.0}{#1}}}

\usepackage{listings} %

\definecolor{commentgreen}{rgb}{0,0.6,0}

\floatstyle{ruled}
\newfloat{listing}{tb}{lst}{}
\floatname{listing}{Listing}

\begin{document}

\title{PointDreamer: Zero-shot 3D Textured Mesh Reconstruction from Colored Point Cloud %
}

\author{Qiao~Yu,
        Xianzhi~Li\textsuperscript{\Letter},
        Yuan~Tang,
        Xu~Han,
        Jinfeng~Xu,
        Long~Hu,
        and~Min~Chen,~\IEEEmembership{Fellow,~IEEE}%
\IEEEcompsocitemizethanks{
\IEEEcompsocthanksitem Qiao Yu, Yuan Tang, Xu Han, Jinfeng Xu are with Huazhong University of Science and Technology, Wuhan, China. (
e-mail: qiaoyu\_epic@hust.edu.cn,  yuantang96@foxmail.com, xhanxu@hust.edu.cn, jinfengxu@hust.edu.cn.\protect)
\IEEEcompsocthanksitem   Xianzhi Li and Long Hu are with School of Computer Science and Technology, Huazhong University of Science and Technology and also with Guangdong HUST Industrial Technology Research Institute, (
e-mail: xzli@hust.edu.cn, hulong@hust.edu.cn.\protect

\IEEEcompsocthanksitem Min Chen is with the School of Computer Science and Engineering, South China University of Technology, Guangzhou, China, and also with the Pazhou Laboratory, Guangzhou, China, E-mail: minchen@ieee.org. 

\IEEEcompsocthanksitem This work was supported by the China National Natural Science Foundation No. 62202182, and in part by China National Natural Science Foundation under Grant 62176101, Grant 62276109, Grant 62322205, and Grant 62272177, in part by the Interdisciplinary Research Program of HUST under Grant 2024JCYJ029, in part by the Department of Education of Guangdong Province under Grant 2021KQNCX157, in part by Guangdong Basic and Applied Basic Research Foundation under Grant 2024A1515030017, Grant 2024A1515011153, Grant 2024A1515010224, and Grant 2024A1515110155, in part by the Postdoctoral Fellowship Program of the China Postdoctoral Science Foundation (CPSF) under Grant GZB20240244, in part by China Postdoctoral Science Foundation under Grant 2024M761016, and in part by Wuhan Natural Science Foundation Exploratory Program (Chenguang Program) under Grant 2024040801020212, and inpart by the Major Research Project of the National Social Science Foundation of China (Grant No. 23\&ZD215). .

\IEEEcompsocthanksitem 
Corresponding author: Xianzhi Li \\
}%
}
\markboth{IEEE Transactions on visualization and computer graphics}%
{Shell \MakeLowercase{\textit{et al.}}: A Sample Article Using IEEEtran.cls for IEEE Journals}

\IEEEtitleabstractindextext{%
\begin{abstract}

\yq{Faithfully reconstructing textured meshes is crucial for many applications. Compared to text or image modalities,  leveraging 3D colored point clouds as input (colored-PC-to-mesh) offers inherent advantages in comprehensively and precisely replicating the target object's 360° characteristics. 
While most existing colored-PC-to-mesh methods suffer from blurry textures or require hard-to-acquire 3D training data, we propose PointDreamer, a novel framework that harnesses 2D diffusion prior for superior texture quality. Crucially, unlike prior 2D-diffusion-for-3D works driven by text or image inputs, PointDreamer successfully adapts 2D diffusion models to 3D point cloud data by a novel project-inpaint-unproject pipeline. Specifically, it first projects the point cloud into sparse 2D images and then performs diffusion-based inpainting. After that, diverging from most existing 3D reconstruction or generation approaches that predict texture in 3D/UV space thus often yielding blurry texture, PointDreamer achieves high-quality texture by directly unprojecting the inpainted 2D images to the 3D mesh. Furthermore, we identify for the first time a typical kind of unprojection artifact appearing in occlusion borders, which is common in other multiview-image-to-3D pipelines but less-explored. To address this, we propose a novel solution named the Non-Border-First (NBF) unprojection strategy. Extensive qualitative and quantitative experiments on various synthetic and real-scanned datasets demonstrate that PointDreamer, though zero-shot, exhibits SoTA performance ( 30\% improvement on LPIPS score from 0.118 to 0.068), and is robust to noisy, sparse, or even incomplete input data.}  Code at: https://github.com/YuQiao0303/PointDreamer.

\end{abstract}

}
\maketitle

 \section{Introduction}

\yq{Faithful 3D mesh reconstruction is a fundamental task in computer vision and graphics. It aims to produce accurate digital recreation of the full geometry and appearance of real-world objects. While 3D generation~\cite{poole2022dreamfusion,wang2023prolificdreamer,wang2025crm} or mesh texturing models~\cite{texture,Text2Tex,perla2024easitex} from text or image inputs have seen remarkable advances, they produce results that exhibit only coarse semantic alignment or viewpoint-specific consistency compared to the target object. As a result, they cannot achieve 360° physically accurate reconstructions with high fidelity. In contrast, 3D colored point clouds provide direct measurements of an object's surface geometry and colors from all perspectives.
This essential feature theoretically provides a unique foundation for faithful reconstruction. Consequently, reconstructing coherent and visually appealing 3D textured meshes from sparse and unstructured colored point clouds (\textit{abbr.} colored-PC-to-mesh) is a pivotal task, with extensive applications like digital twin~\cite{digital_twin}, metaverse~\cite{metaverse}, cultural heritage preservation~\cite{wang2024xscalenvs}, etc.} %

\begin{figure}
  \includegraphics[width=0.99\linewidth]{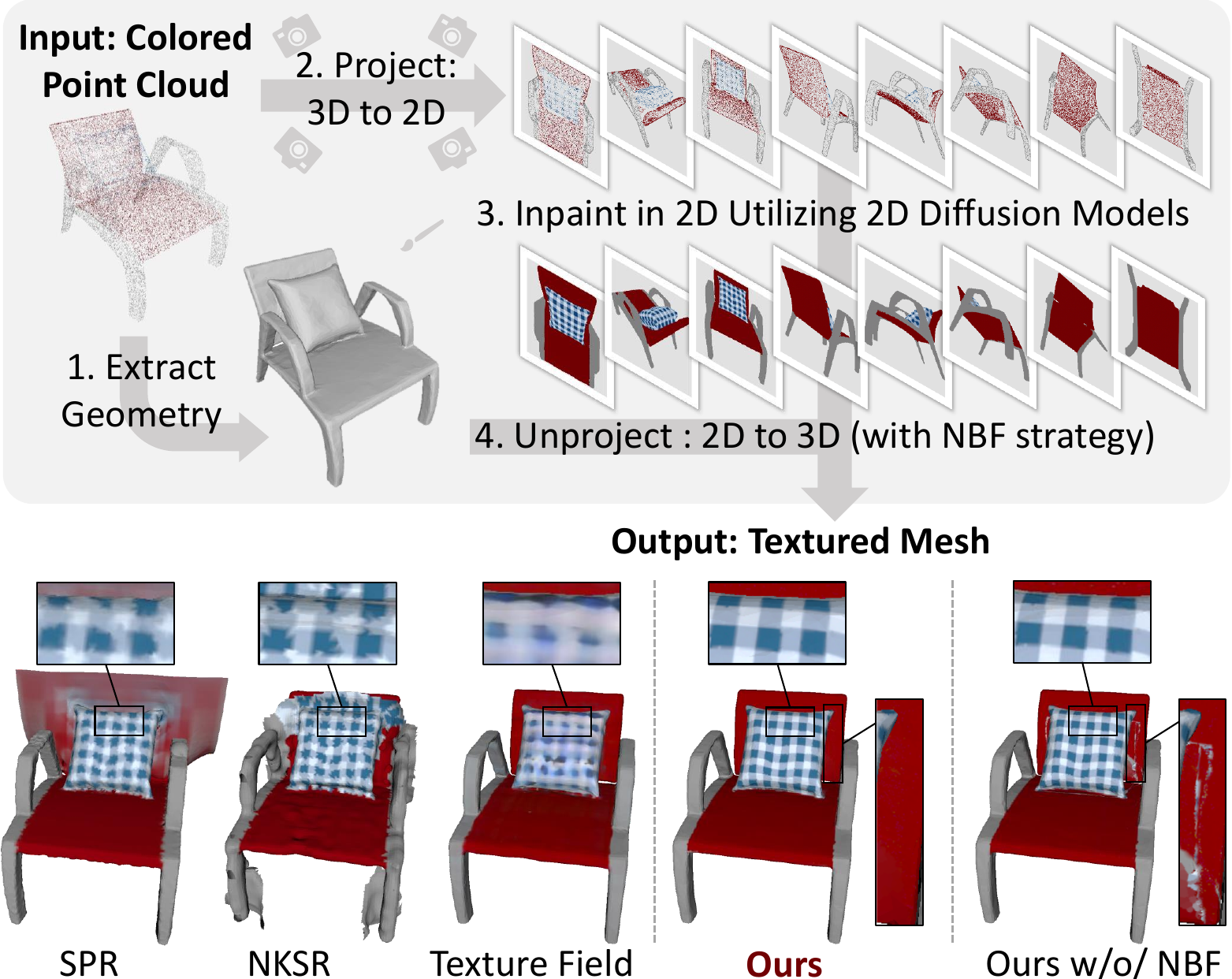}
  \caption{PointDreamer is a zero-shot framework to reconstruct 3D textured meshes from colored point clouds. It produces textures with higher quality compared to baseline methods. Its specially designed NBF unprojection strategy effectively addresses the border-area artifact issue. %
  }
  
  \label{fig:teaser}
\end{figure}

\yq{Different from 3D generation based on text or images, point cloud reconstruction faces its own challenges.} While recent point cloud reconstruction methods~\cite{Boulch_2022_POCO,ALTO,saconvonet} have achieved relatively satisfactory geometry reconstruction quality, color reconstruction remains challenging, especially for low-density point clouds. Unlike images that provide dense pixel-level representations, the inherent sparse nature of point clouds makes it difficult to reconstruct dense colors. %
Regarding this, existing methods either blend input points' colors by interpolation~\cite{kazhdan2013screened} or overfitting~\cite{Wei_2021_DHSP3D,huang2023nksr}, or train color prediction networks~\cite{gupta20233dgen,ColorMesh} with 3D datasets. However, they often yield blurring-looking textures; see SPR~\cite{kazhdan2013screened}, NKSR~\cite{huang2023nksr}, and Texture Field~\cite{OechsleTextureField,gupta20233dgen} in Figure~\ref{fig:teaser}. Also, 3D training data are notoriously challenging to acquire.

While the colored-PC-to-mesh task remains challenging, 2D image generation has recently thrived with impressive plausibility, level of detail, and generalization capability. %
Therefore, many works~\cite{poole2022dreamfusion,liu2023syncdreamer} leverage 2D diffusion models for 3D generation. Since the scale of existing 2D datasets far exceeds that of 3D datasets, leveraging diffusion models pre-trained on extensive 2D datasets can also significantly alleviate the reliance on 3D datasets. This inspires us to adopt 2D diffusion models for the colored-PC-to-mesh task, to address the low-quality color reconstruction issue.

However, it is not trivial to implement this idea: \textbf{(1)} \yq{Existing 2D diffusion models~\cite{latent_diffusion, controlnet} are predominantly developed to be conditioned on text or image inputs, fundamentally incompatible with point cloud data. Consequently, existing 2D-diffusion-based 3D generation methods are driven typically by text~\cite{lin2023magic3d,Chen_Fantasia3D} or images~\cite{wang2025crm,tang2024lgm}, leaving point-cloud-based reconstruction under-explored. }
\textbf{(2)} Existing 2D-diffusion-for-3D  methods often yield blurry appearances~\cite{tang2023dreamgaussian} by predicting colors in 3D space as a texture field~\cite{liu2023one2345,Chen_Fantasia3D}, \yq{a radiance field~\cite{melaskyriazi2023realfusion,liu2023zero1to3,Tang_makeIt3D}} or 3D Gaussians~\cite{tang2023dreamgaussian,tang2024lgm}. How to leverage 2D reference images to generate high-quality textures requires more attention. \textbf{(3)} Intermediate 2D multiview images generated by diffusion models often have inconsistencies with each other or with the reconstructed 3D geometry at occlusion border areas. When one part of an object occludes another, the generated 2D images and 3D geometry should predict highly consistent occlusion borders. Failing to do so would cause artifacts; see ``Ours w/o/ NBF" in Figure~\ref{fig:teaser}, with significant artifacts near the border of the pillow and the chair.

We address these challenges one by one. \textbf{First}, to bridge the gap between 2D diffusion models and 3D point clouds, our intuition is that, reconstructing 3D textured meshes from colored point clouds is analogous to inpainting 2D sparse images by filling empty pixels. They both aim to somehow \emph{dreaming} the missing areas of the sparse input, to achieve a more complete and coherent representation. 
Luckily, 2D diffusion-based inpainting methods~\cite{DDNM} excel even with mask ratios over 80\%, making them robust to sparse input points.  Trained on large-scale 2D datasets, these models generalize well across domains and require no additional training or fine-tuning, enabling zero-shot point cloud reconstruction. 
\textbf{Second}, to address the blurring issue, we propose to directly unproject the generated multiview images onto the 3D mesh, instead of predicting colors in 3D~\cite{gupta20233dgen,huang2023nksr} or UV~\cite{Wei_2021_DHSP3D} spaces like existing methods. \textbf{Third}, to address the occlusion-border inconsistency issue, we design a ``Non-Border-First" (NBF) unprojection strategy. It detected border areas by leveraging the correspondence of occlusion border in 2D image space and invisibility border in UV space, and then prioritizes non-border views' images during unprojection, thus avoiding artifacts. %

Putting everything together, we propose \emph{PointDreamer}, a novel zero-shot framework to reconstruct high-quality textured meshes from colored point clouds, as shown in Figure~\ref{fig:teaser}. First, we extract geometry (untextured mesh) from the point cloud by an existing method POCO~\cite{Boulch_2022_POCO}. Second, we project the point cloud into 2D space from a fixed set of viewpoints, producing multiview sparse images. Third, we inpaint the empty pixels with an off-the-shelf 2D diffusion model, forming dense images. Finally, we directly unproject the 2D images onto the mesh. Unlike most existing methods that predict colors in 3D~\cite{gupta20233dgen,ColorMesh} or UV spaces~\cite{Wei_2021_DHSP3D}, directly adopting the colors of 2D diffusion models' output high-quality 2D images leads to clear textures. In particular, our designed NBF unprojection strategy effectively avoids artifacts caused by inconsistencies in occlusion border regions. 

Overall, we list our contributions below:

\begin{itemize}
\yq{
\item We propose PointDreamer, a SoTA framework for 3D textured mesh reconstruction from colored point cloud, with multiple advantages: high-quality texture, zero-shot, and robust to noisy, sparse, or even incomplete input data.

\item Existing 2D diffusion models are predominantly developed for text or image inputs, thus confining existing 2D-diffusion-for-3D methods to these inputs too. Unlike them, PointDreamer resolves the intrinsic incompatibility between 2D diffusion and colored 3D point clouds, by a novel project-inpaint-unproject pipeline.

\item Many existing 3D reconstruction or generation methods yield blurry texture by predicting colors in 3D or UV space. In contrast, PointDreamer achieves high-quality texture by directly unprojecting predicted 2D images to 3D.

\item We identify for the first time a common but less-explored artifact in multiview-image-to-3D pipelines caused by inconsistent prediction near occlusion borders. We propose a novel strategy named Non-Border-First (NBF) unprojection to effectively address this issue.
}

\end{itemize}
Experiments on various benchmarks show the \textbf{SoTA} performance of PointDreamer, by significantly outperforming baseline methods quantitatively and qualitatively. 

\section{Related Work}

\subsection{Textured Mesh Reconstruction from Colored Point Cloud}
Existing methods can be categorized based on their reliance on training data. 
\textbf{Non-data-driven} methods rely solely on the input point cloud. For instance, (Screened) Poisson Surface Reconstruction (PSR, SPSR)~\cite{kazhdan2006poisson,kazhdan2013screened} reconstructs geometry by solving Poisson equations, and many 3D processing tools~\cite{cignoni2008meshlab,open3d} extend it with color support via linearly blending input point colors. DHSP3D~\cite{Wei_2021_DHSP3D} iteratively optimizes a MeshCNN in 3D space and XYZ/RGB maps in UV space by self-supervision. 
On the other hand, \textbf{data-driven} methods train neural networks on 3D datasets to generate textured meshes. For example, 3DGen~\cite{gupta20233dgen} trains a triplane variational autoencoder to predict signed distance field and color values for each 3D tetrahedra grid vertex, and extracts textured meshes by marching tetrahedra~\cite{shen2021dmtet}. \textbf{Hybrid} methods employ different strategies for geometry and texture reconstruction. 
NKSR~\cite{huang2023nksr} adopts data-driven geometry reconstruction by learning neural kernel fields, while using non-data-driven color reconstruction (in its supplementary file), by optimizing a 3D textured field~\cite{OechsleTextureField} to overfit the input points' colors. ColorMesh~\cite{ColorMesh} employs non-data-driven geometry reconstruction via graph cuts, paired with a data-driven texture network to inpaint colors in 2D image space. However, its texture network is an encoder-decoder CNN trained with limited data. This may limit its performance and generalization capability, compared to our adopted zero-shot 2D-diffusion-based inpainting approach trained with extensive 2D data. %

\subsection{\yq{2D Diffusion for 3D Generation and Texturing}}

\begin{figure}
    \centering
    \includegraphics[width=1\linewidth]{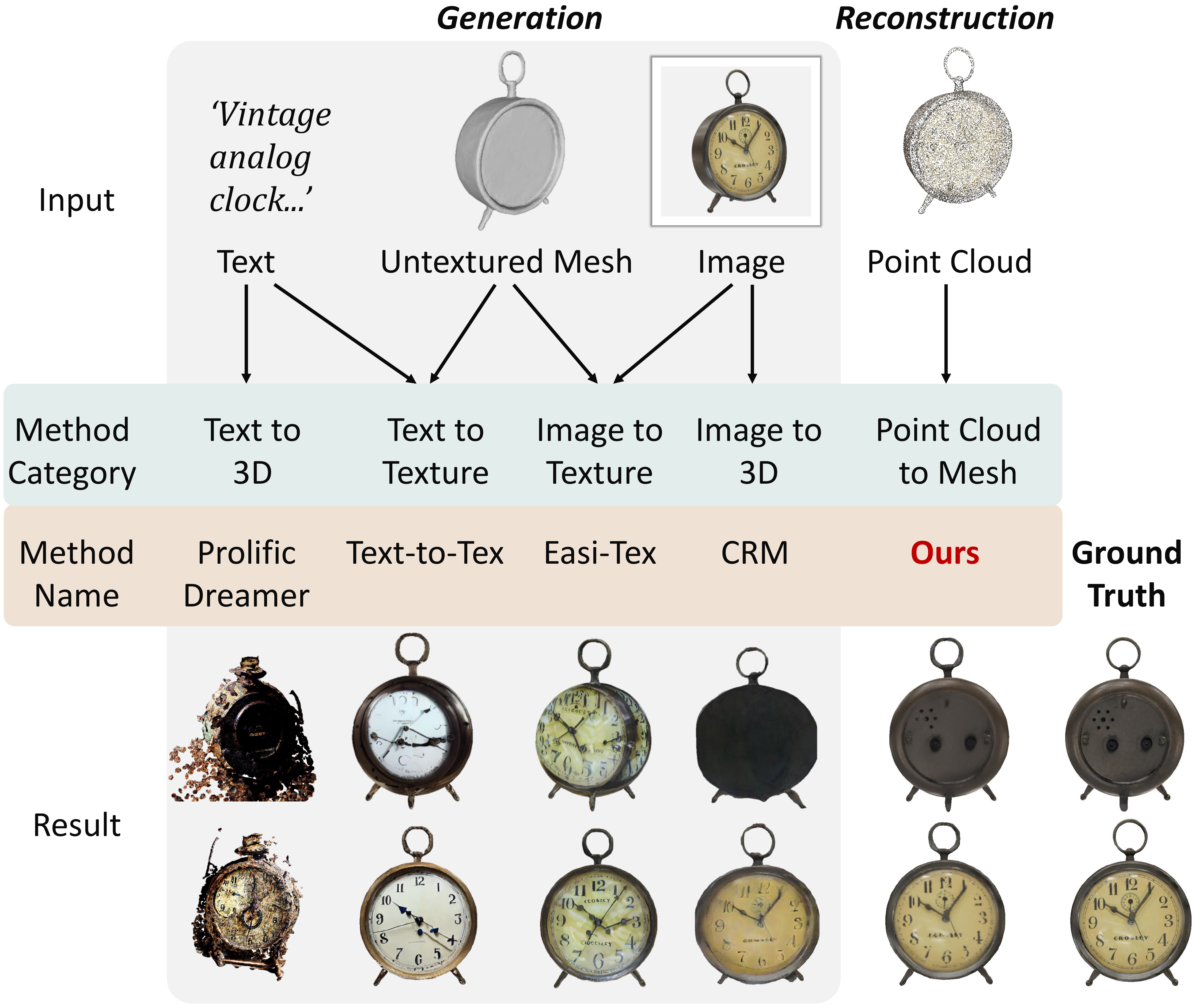}
    \caption{\yq{Comparison with different 2D-diffusion-for-3D methods. }}
    \label{fig:gen_illustration}
\end{figure}

\yq{With the success of 2D diffusion models, researchers have explored utilizing them for 3D generation and 3D mesh texturing.}
Most of these generation works optimize or learn a NeRF~\cite{liu2023zero1to3,Tang_makeIt3D}, a DMTet with texture field~\cite{wang2023prolificdreamer, qian2023magic123,Chen_Fantasia3D}, or 3D Gaussianss by Score Distillation Sampling~\cite{poole2022dreamfusion} or reconstruction loss to make the generated 3D scene similar to the result of a given 2D diffusion model. 
View-conditioned diffusion models~\cite{liu2023zero1to3} or multi-view diffusion models~\cite{szymanowicz23viewset_diffusion,liu2023syncdreamer,shi2023MVDream} are developed to produce multiview-consistent images for subsequent 3D generation or texturing. \yq{Later, Large Reconstruction Models~\cite{hong2023lrm,sf3d2024,wang2025crm} are proposed, demonstrating the effectiveness of directly using feed-forward networks for 3D generation. In addition, inpainting-based methods are popular in the field of generating texture for 3D meshes~\cite{texture, Text2Tex, perla2024easitex}. They use geometry-aware diffusion models to progressively inpaint the complete texture. However, conditioned on only text or a single image instead of a point cloud, the above methods align more closely with ``generation" instead of ``reconstruction". In other words, they theoretically cannot faithfully reconstruct the target object; see Figure~\ref{fig:gen_illustration} for an example. Also, they may generate blurry textures as shown in ``CRM'' in Figure~\ref{fig:gen_illustration}, and may suffer from artifacts in occlusion border areas. }

\subsection{Border-Area-Inconsistency Issue in 3D Texturing}
Mesh texture reconstruction based on real-scanned or AI-generated multiview images is widely used in applications such as 3D reconstruction and generation~\cite{long2023wonder3d}. Due to scanning errors and imperfections in generative models, inconsistencies often arise between different views' images or between images and the mesh, especially around occlusion borders where one part of the object occludes another. However, this issue remains largely unexplored. 
Most texture mapping methods~\cite{huang2020adversarial,fu2021seamless,fu2020joint,waechter2014let} design complex optimization objectives and strategies for camera extrinsics, etc. They typically focus on issues like blurring, ghosting, and seams, leaving border-area-inconsistency unexplored. Also, they predominantly rely on RGBD data, and thus are unsuitable for scenarios without dense depth data like point cloud reconstruction. On the other hand, most 3D generation approaches directly fuse multiview images' colors for mesh texturing. Specifically, they use these images as supervision during optimization~\cite{long2023wonder3d}, or directly compute image colors' weighted sum as the mesh's colors~\cite{wu2024unique3d}. To address the border-area-inconsistency issue, we design a Non-Border-First unprojection strategy, and it can be used in any method that unprojects multiview images to a mesh.

\section{Method}
Given a 3D point cloud with per-point XYZ coordinates and RGB colors, our goal is to reconstruct its associated textured mesh. To avoid the blurring effect of existing works, we predict colors in 2D space by diffusion-based 2D inpainting, leveraging the powerful 2D diffusion priors.
Figure~\ref{fig:teaser} presents our designed pipeline with four steps.
(1) We employ a geometry extraction module to reconstruct an untextured mesh from the input point cloud. (2) With a set of fixed viewpoints, we perform 3D-to-2D conversion by
projecting the input point cloud into 2D, producing sparse multi-view images.
(3) We conduct color prediction in 2D space by inpainting the empty pixels in the sparse images to form dense ones based on a pre-trained 2D diffusion model~\cite{guided_diffusion}. (4) We propose a novel Non-Border-First strategy to convert the 2D results back to 3D by unprojecting the colors in the dense images to the untextured mesh to produce the desired textured mesh. 

Note that with the 2D diffusion priors, our method is zero-shot requiring no extra training.
It takes only about 61s for PointDreamer to reconstruct a shape of 30k points on an NVIDIA A100 GPU, compared to other 2D-diffusion-for-3D methods such as 
Zero-1-to-3~\cite{liu2023zero1to3} (\textasciitilde 20 m), DreamGaussin~\cite{tang2023dreamgaussian} (\textasciitilde 2 m), Wonder3D~\cite{long2023wonder3d} (2\textasciitilde 3 m), LGM~\cite{tang2024lgm} (\textasciitilde 65 s), \yq{Texture~\cite{texture} (\textasciitilde 5 m), Text2Tex~\cite{Text2Tex} (\textasciitilde 15 m), Easi-Tex~\cite{perla2024easitex} (\textasciitilde 6 m), and ProlificDreamer~\cite{wang2023prolificdreamer} (several hours)}.

\begin{figure}[t]
    \centering
\includegraphics[width=1.0\linewidth]{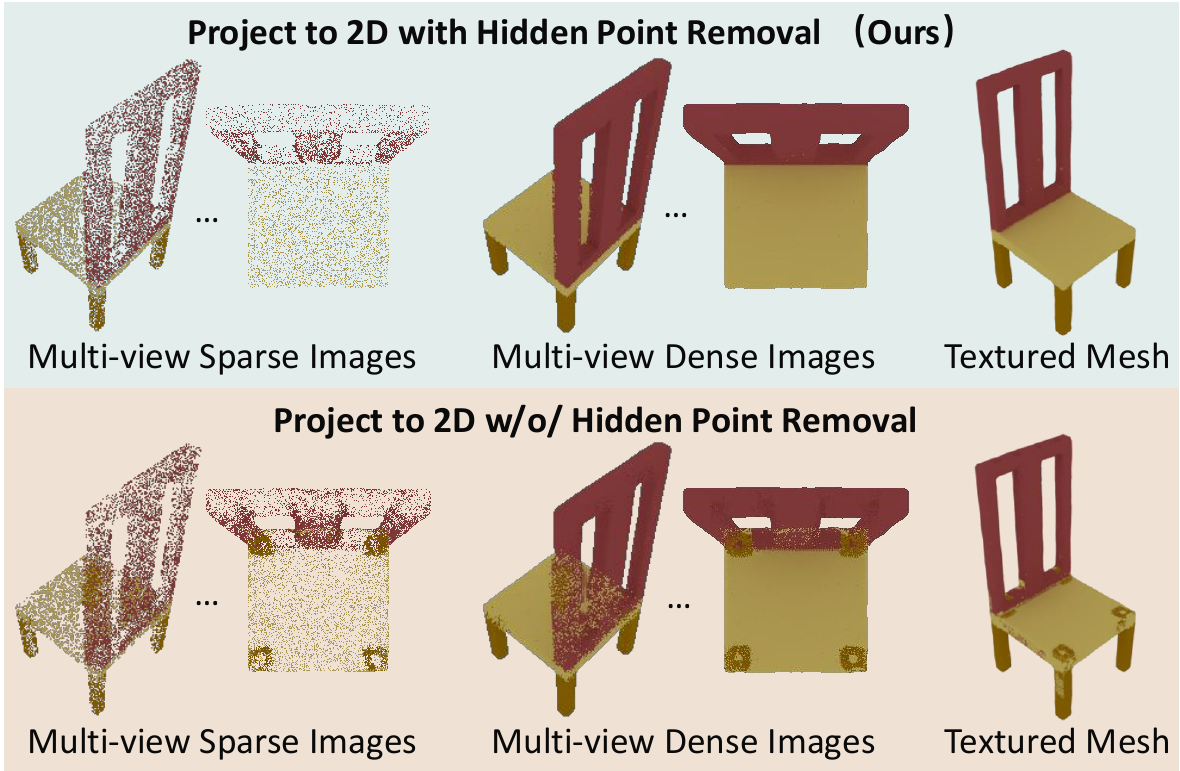}
    \caption{Hidden point removal aims \yq{to avoid significant stippled artifacts}} 
    \label{fig:hidden}
\end{figure}

\begin{figure*}[t]
  \centering
  \includegraphics[width=0.78\linewidth]{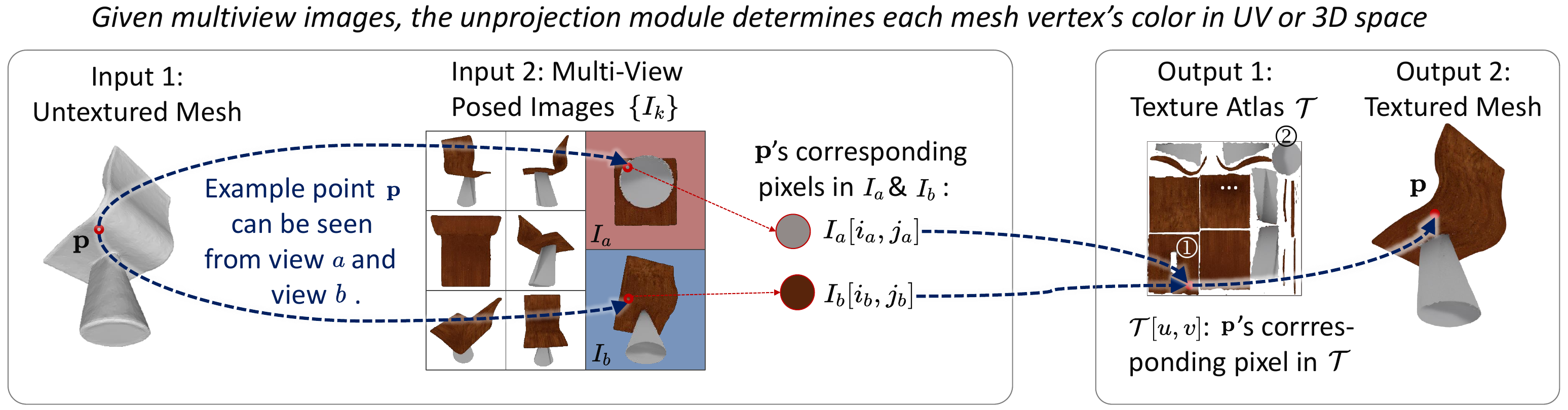}
  \caption{
 \yq{ The unprojection module takes an untextured mesh and corresponding multiview images as input, to output the associated textured mesh.}
  }
  \label{fig:unproject}
\end{figure*}

  \begin{figure*}[t]
  \centering
  \includegraphics[width=1.0\linewidth]{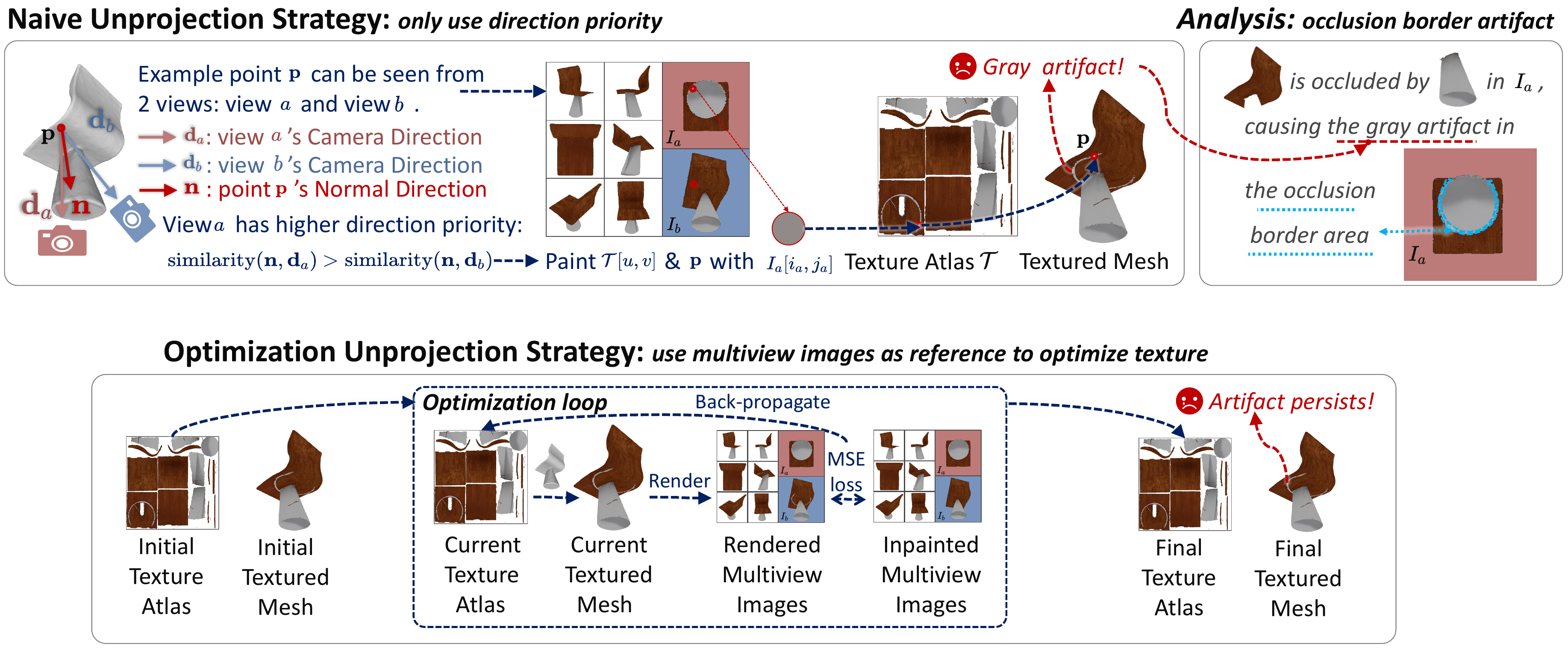}
  \caption{
 \yq{ \textbf{The naive unprojection strategy} chooses one best view for each texture atlas pixel by considering only the direction priority (similarity between the 3D point normal direction and the camera direction).  This can lead to border-area artifacts. \textbf{The optimization-based unprojection strategy} uses multiview images as a reference to optimize the texture atlas with per-pixel MSE loss. However, the artifacts still exist.}
  }
  \label{fig:naive_opt}
\end{figure*}

 \begin{figure}[t]
  \centering
  \includegraphics[width=1.0\linewidth]{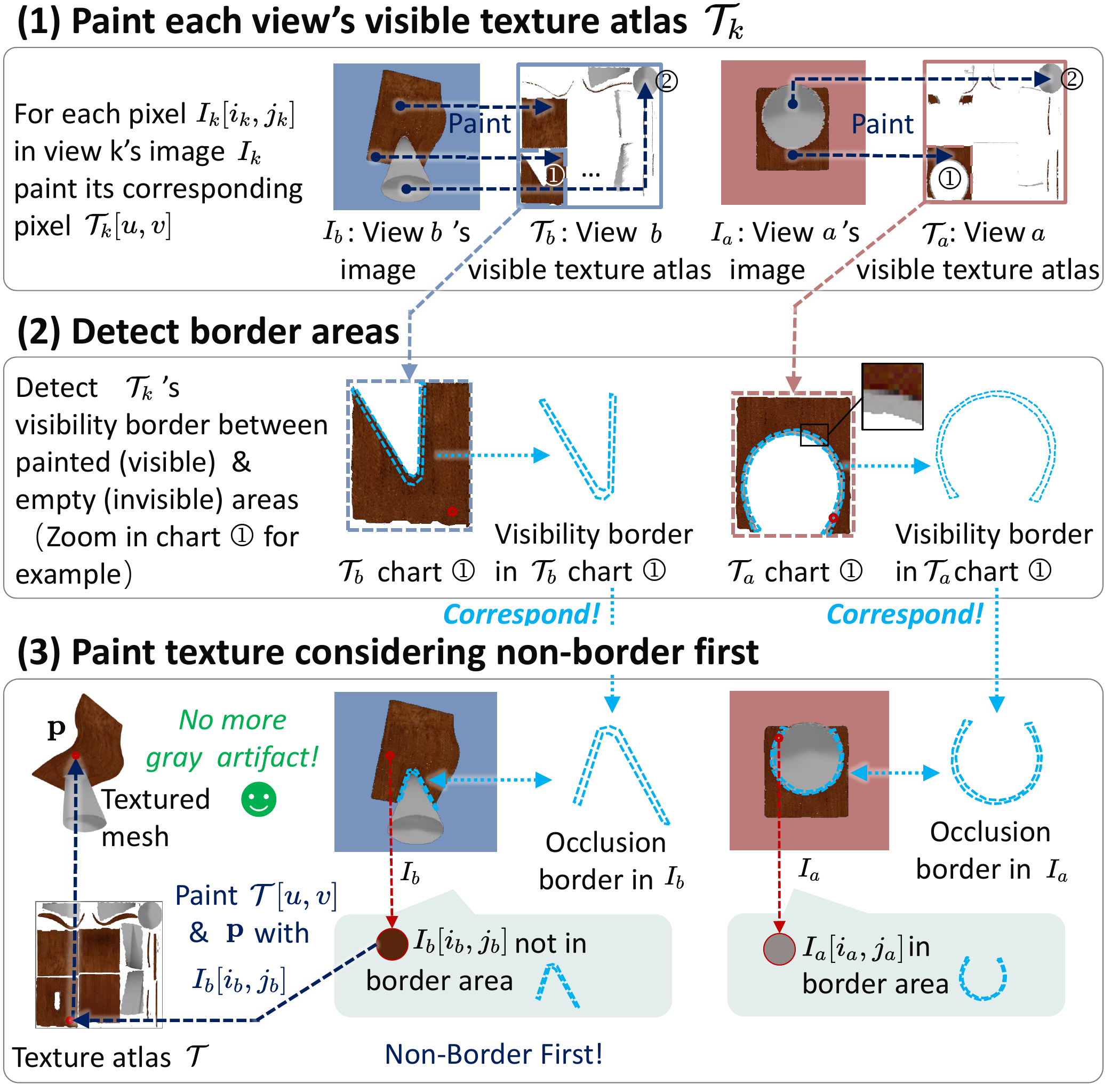}
  \caption{
\yq{  Illustration of the proposed NBF unprojection strategy. By painting each view's visible texture atlas, we can detect its visibility border areas, which correspond to occlusion borders in the corresponding view's 2D image. Then paint the texture by prioritizing non-border areas.}
  }
  \label{fig:NBF_method}
\end{figure}

\subsection{Geometry Extraction: Point to Surface}
The first step in PointDreamer is to reconstruct an untextured mesh from the input point cloud.
Since color information does not need to be considered in this step, many existing point-to-surface methods~\cite{onet,convonet,shen2021dmtet} can be employed. 
In our implementation, we directly adopt the state-of-the-art POCO\footnote{https://github.com/valeoai/POCO}~\cite{Boulch_2022_POCO} for its high performance. 
We conduct experiments comparing different geometry reconstruction methods in Section~\ref{sec:exp} and our supplementary file.

\subsection{Projection: 3D to 2D}
Directly projecting the whole point cloud into 2D space would incorporate points from occluded parts that should not be visible from the given view. This would \yq{cause significant stippled artifacts} in the inpainting results and subsequent mesh textures; see Figure~\ref{fig:hidden} for an example. 
Therefore, before generating the sparse images, we process the input point cloud by the ``Hidden Point Removal'' operator~\cite{hiddenPointRemoval}, which works by transforming the input and extracting the points that reside on its convex hull. Besides, we compare the depth values of points and the extracted untextured mesh to further remove some invisible points. 
With pre-set camera parameters of $K$ viewpoints ($K=8$ in our implementation) and associated visible point clouds, we conduct camera transformation to these points to get their corresponding pixel coordinates in 2D image space. These pixels are painted according to their associated 3D points' colors, producing $K$ sparse images. 
\textbf{We compare different $K$ values in our supplementary file.}

\subsection{Inpainting: Sparse to Dense}
Any 2D inpainting method that fills empty pixels in input images can serve as our inpainting module. We propose to use the state-or-the-art DDNM\footnote{https://github.com/wyhuai/DDNM}
~\cite{DDNM} based on a pre-trained unconditional 2D diffusion model\footnote{https://github.com/openai/guided-diffusion}
~\cite{guided_diffusion} to inpaint the multi-view sparse images 
into dense ones $\mathbf{I} = \{I_k\}_{k=1}^K$. %
In this way, the 2D diffusion model's strong prior can facilitate high-quality inpainting. So far, we have completed color prediction purely in 2D space instead of 3D or UV space.

\subsection{Unprojection: 2D to 3D}
With color prediction conducted in 2D space, we now need to convert the result back to 3D. Specifically, the goal is to use the extracted untextured mesh and inpainted multi-view posed images, to generate the associated textured mesh. Regarding this, we design a novel approach namely ``Non-Border-First Unprojection''.

Figure~\ref{fig:unproject} illustrates the basic idea of the unprojection module. Specifically, we represent mesh texture as a texture atlas $\mathcal{T}$ by applying UV mapping to a mesh $\mathcal{M}$. We follow existing works~\cite{gao2022get3d,long2023wonder3d} to conduct UV mapping by Xatlas~\cite{young2022xatlas}. A texture atlas is an RGB image containing different charts \yq{(e.g. \ding{172}, \ding{173} ... in Figure~\ref{fig:unproject}). Each chart is a continuous area in UV space that corresponds to a continuous segment of $\mathcal{M}$ in 3D space. For example, in Figure~\ref{fig:unproject}, chart \ding{172} corresponds to the bottom side of the brown seat, and chart \ding{173} corresponds to the bottom side of the gray cone.} 
Each pixel $\mathcal{T}[u,v]$ of a chart corresponds to a surface point $\mathbf{p}$ of $\mathcal{M}$. Therefore, our goal is to assign a color for each chart pixel $\mathcal{T}[u,v]$.
Given a set of multi-view posed images $\mathbf{I} = \{I_k\}_{k=1}^K$, $\mathbf{p}$ can be seen in zero, one, or more images. If $\mathbf{p}$ is visible in $I_k$, we denote $I_k$'s corresponding pixel as $I_k[i_k,j_k]$.
See the top of Figure~\ref{fig:unproject} as an example, where a 3D point $\mathbf{p}$ is visible in two views $a$ and $b$, and we mark it with a red circle together with its corresponding multi-view image pixels $I_a[i_a,j_a]$, $I_b[i_b,j_b]$, and texture atlas pixel $\mathcal{T}[u,v]$. 

Point $\mathbf{p}$'s visibility in view $k$ can be calculated by whether its depth value is no bigger than the corresponding pixel of the $k$'s depth map. If $\mathbf{p}$ is visible only in one view, the corresponding color of $I_k[i_k,j_k]$ can be directly adopted as the color of $\mathbf{p}$ and $\mathcal{T}[u,v]$. If $\mathbf{p}$ can be seen in no view at all, we can set some rules to deal with it. If $\mathbf{p}$ can be seen in more than one view, we can try to fuse the corresponding colors or select one best view.
Various unprojection strategies differ in how to fuse views or select the best view. Here we introduce three different implementations for unprojection.

\subsubsection{Naive Unprojection: Only Use Direction Priority}
An intuitive idea to pick the best view for $\mathbf{p}$ or $\mathcal{T}[u,v]$ is by direction priority, as shown in `Naive Unprojection Strategy' on top of Figure~\ref{fig:naive_opt}.
Specifically, for each view $k$ where point $\mathbf{p}$ can be seen, we calculate the direction priority score $s_\mathbf{p}(k)$, defined as the cosine similarity of $\mathbf{p}$'s normal direction $\mathbf{n}$ and view $k$'s camera direction $\mathbf{d}_k$. 
The view with the highest direction priority is chosen to paint $\mathbf{p}$ and $\mathcal{T}[u,v]$.
For example, in the top row of Figure~\ref{fig:naive_opt}, $\mathbf{p}$ is painted as gray according to $I_a[i_a,j_a]$ since
view $a$ has a higher direction priority.
Also, if there are a few points invisible from any view, we can still assign a view by direction priority.

While this naive unprojection strategy serves adequately for most points, it produces artifacts under certain circumstances; see again the top row of Figure~\ref{fig:naive_opt}, where a gray artifact is visible in the reconstructed textured mesh that should have been brown. 
As shown in `Analysis' in Figure~\ref{fig:naive_opt}, such artifacts appear in \emph{border areas}: the area near the border of the occluded and occluding part of the shape when viewing from view $k$. For example, in view $a$, the brown seat's bottom side is occluded by the gray bottom of the chair. The area near the edge of the brown and gray parts is a border area.
In such areas, the inpainting module and the geometry extraction module may inconsistently delineate the boundary separating the two parts. For example, the gray bottom is larger in the inpainted image $I_a$ than of the reconstructed mesh, thus resulting in the artifact.

\subsubsection{Optimization-based Unprojection}
Instead of selecting one single view by direction priority, 3D generation method DreamGaussian~\cite{tang2023dreamgaussian} adopts an optimization-based strategy to get the textured mesh from multi-view images. \yq{As shown in `Optimization Unprojection Strategy' of Figure~\ref{fig:naive_opt}, the idea is to optimize the texture atlas by a per-pixel loss between the inpainted multi-view images $\{I_k\}_{k=1}^K$ and the rendered images of the reconstructed textured mesh.} In this way, the optimization process \emph{fuses} corresponding colors in different views instead of selecting the best one. However, it cannot ignore the inconsistent colors in the \emph{border areas}, so the artifacts can only be reduced instead of eliminated. This inspires us to design a non-border-first strategy to address this issue.

\subsubsection{\yq{Non-Border-First Unprojection (NBF)}}

\yq{
Building upon the above analysis, we propose the \emph{Non-Border-First (NBF)} Unprojection strategy to eliminate artifacts arising from inconsistent predictions in border areas. The core principle is to prioritize assigning colors from non-border regions across views when filling the texture atlas. 

\para{Border Area Detection.} 
To prioritize non-border areas, we first need to detect border areas in each view. Since the direct detection of occlusion borders within multiview 2D images is challenging, we propose to shift the detection process to UV space. This approach leverages a key correspondence we observe empirically: occlusion borders in image $I_k$ (2D space) often correspond to visibility borders in ``view $k$ visible texture atlas $\mathcal{T}_k$'' (UV space), where $\mathcal{T}_k$ is defined as a texture atlas in which only visible areas in view $k$ are filled and rest invisible areas are empty. 

For example, Figure~\ref{fig:NBF_method} (1) illustrates view $a$ and view $b$'s visible texture atlas $\mathcal{T}_a$ and $\mathcal{T}_b$, respectively. As the navy blue dashed arrows show, the bottom side of the brown seat in $I_a$ and $I_b$ are mainly used to paint corresponding pixels in chart \ding{172} of $\mathcal{T}_a$ and $\mathcal{T}_b$, and the bottom side of the gray cone is mainly used to paint chart \ding{173}.
Zooming in, we can see in Figure~\ref{fig:NBF_method} (2) that, the representive chart \ding{172} in both $\mathcal{T}_a$ and $\mathcal{T}_b$ can be divided into a brown area (visible) and an empty area (invisible), respectively. We mark their border areas in sky blue dashed lines and denote such borders as ``visibility border''. Note that the brown area corresponds to the bottom side of the brown seat of the chair, and the empty area corresponds to the parts that are occluded by the gray cone, as we marked by arrows with text ``Correspond!" in sky blue in Figure~\ref{fig:NBF_method}. In other words, the invisible/visible areas in $\mathcal{T}_k$ correspond to the occluded/occluding areas in $I_k$. The reason for this correspondence is that, a continuous area within a chart of the texture atlas usually corresponds to a continuous segment of the 3D surface. When a chart contains both visible and invisible regions in a given view $k$, it often indicates that the corresponding 3D areas of invisible regions are occluded by some other part of the shape.

\para{Border-Priority-Based Unprojection.} 
Exploiting this correspondence, visibility borders in $\mathcal{T}_k$ (detected via edge detection algorithms) serve as proxies for occlusion borders in
$I_k$. Consequently, NBF prioritizes non-border regions during texturing. For instance, in Figure~\ref{fig:NBF_method} (3), example 3D point $\mathbf{p}$ and its corresponding pixel $I_a[i_a,j_a]$ and $I_b[i_b,j_b]$ in $I_a$ and $I_b$ are marked in small red circles. Since $I_a[i_a,j_a]$ and $I_b[i_b,j_b]$ are in border and non-border areas, respectively, the non-border-first principle guides us to pritize $I_b[i_b,j_b]$ and paint point $\mathbf{p}$ as $I_b[i_b,j_b]$'s color (brown). In this way, we effectively eliminate the gray artifact in naive or optimization unprojection caused by inconsistent prediction in border areas. Putting everything together, the detailed steps of the proposed NBF unprojection strategy are as follows. 

}
\begin{enumerate}
\item We paint ``view $k$ visible texture atlas map'' $\mathcal{T}_k$ for each view $k$ defined as above.    

\item We calculate the border areas by dilating the edges between visible and invisible areas in $\mathcal{T}_k$ given a certain dilation kernel size as a hyperparameter.

\item We paint the final texture atlas $\mathcal{T}$ considering only non-border areas in ${\mathcal{T}_k}$. During this, if a point is visible in more than one view's non-border areas, we select the view with the highest direction priority.

\item If there remain some unpainted pixels in $\mathcal{T}$, check the previously ignored border areas to paint them, during which we still select among views by direction priority.

\item If some points cannot be seen from either view, we assign each of them with a view by direction priority considering all areas in all views whether they're border areas or not.

\item We experimentally find that additionally optimizing from the generated texture atlas by only calculating the loss of non-border pixels would further enhance the results.

\end{enumerate}

\section{Experiments and Results}
\label{sec:exp}

\subsection{Experimental Settings}

\subsubsection{Datasets}
To evaluate the performance of our method against other competitors, we use three benchmark datasets: %

(i) \textbf{ShapeNetCoreV2}~\cite{shapenet2015}:
A large-scale synthesis 3D object dataset, where we follow~\cite{gao2022get3d} to use the official test splits of \emph{chair}, \emph{car}, and \emph{motorbike} categories for evaluation, since they contain relatively complex textures.
The test sets of the three categories contain around 1300, 690, and 70 samples, respectively.

(ii) \textbf{Google Scanned Objects (GSO)}~\cite{downs2022google}:
A real-scanned 3D object dataset and \textbf{we use all its 1030 samples for evaluation, unlike existing works~\cite{liu2023syncdreamer,long2023wonder3d} that only use 30 of them.}

(iii) \textbf{OmniObject3D}~\cite{wu2023omniobject3d}:
A real-scanned 3D object dataset with 6000 samples. For efficiency, we randomly select 100 objects for evaluation.

\begin{figure}[!ht]
  \centering
  \includegraphics[width=\linewidth]{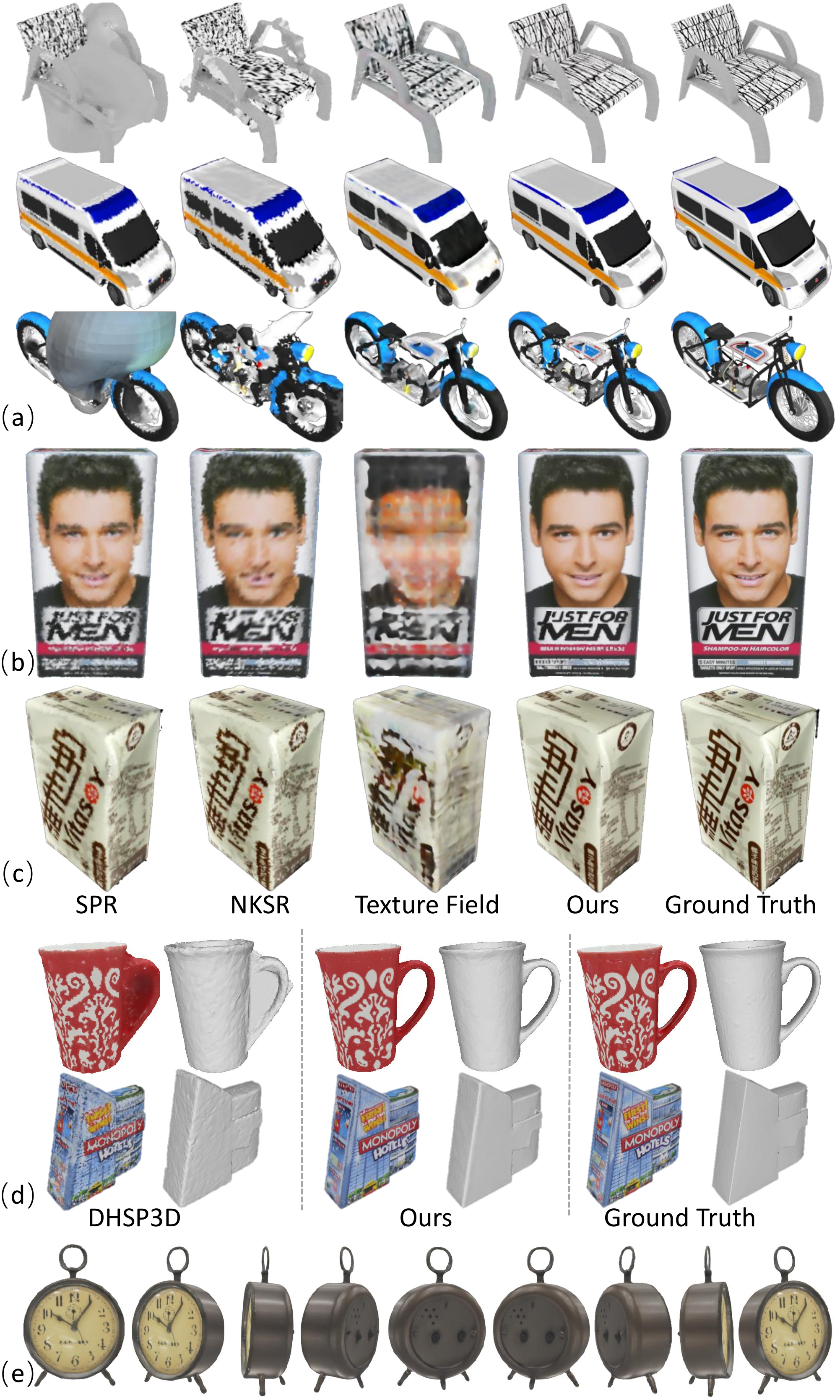}
  \caption{Qualitative results on (a) ShapeNetCoreV2 dataset (Chair, Car, Motorbike Category), (b) GSO dataset, (c) OmniObject3D dataset. (d) Comparison with DHSP on GSO. (e) Our Multiview Consistency. 
  }
  \label{fig:results}
\end{figure}

For each textured mesh from the above datasets, we sample 30k colored points as input. Note that, our approach is zero-shot and can be directly applied to objects from various datasets or categories, requiring no extra training but only an off-the-shelf 2D diffusion model~\cite{guided_diffusion}.

\subsubsection{Metrics}
To evaluate the quality of the reconstructed 3D textured mesh, we follow existing works~\cite{liu2023zero1to3,liu2023syncdreamer,long2023wonder3d,zhang2022meshInversion} to compare the similarity between multi-view images rendered by the reconstructed mesh and ground-truth mesh.
Specifically, we render 20 images per shape for a thorough assessment by distributing cameras on the vertices of a regular icosahedron. Four similarity metrics are calculated:
Peak Signal-to-Noise Ratio (PSNR), Structural Similarity Index Measure (SSIM)~\cite{SSIM}, Learned Perceptual Image Patch Similarity (LPIPS)~\cite{lpips}, and Fréchet Inception Distance (FID) ~\cite{FID}. PSNR and SSIM focus more on pixel-level similarity, while LPIPS and FID imitate human perception better.
Since we focus on texture instead of geometry reconstruction, \textbf{we put geometry evaluation results in our supplementary file.}

\subsubsection{Baselines}
To fully evaluate the performance of our method, we compare it with three kinds of baseline methods:

(1) \emph{Screened Poisson Reconstruction (SPR)~\cite{kazhdan2013screened}}: SPR is a classical mesh reconstruction method. We adopt the implementation of the commonly-used tool MeshLab~\cite{cignoni2008meshlab}, which generates texture by linearly blending colors of the input point cloud. Note that SPR requires per-point normal as input, which we estimate via MeshLab's default Principal Component Analysis (PCA).

(2) \emph{DHSP3D~\cite{Wei_2021_DHSP3D} and NKSR~\cite{huang2023nksr}}: These are two recent approaches for mesh reconstruction which generate textures by overfitting the input points' colors. We employ the official code provided by the authors~\footnote{https://github.com/weixk2015/DHSP3D,\\https://github.com/nv-tlabs/NKSR/blob/public/examples/\\recons\_colored\_mesh.py}.
Similar to SPR, NKSR also requires point normals as additional input.

(3) \emph{Texture Field (T.F.) ~\cite{OechsleTextureField}}:
Since we cannot find any existing open-source texture-field-based method requiring only colored point clouds as input, we thus design a baseline network inspired by~\cite{gupta20233dgen}.
The key idea is to learn a 3D feature tri-plane from the input point cloud and then decode RGB values for the query points. We train it on the official training split of the mentioned three categories of ShapeNetCoreV2, supervised by per-point color MSE loss following~\cite{gupta20233dgen}. 
During inference, we employ the above-mentioned POCO~\cite{Boulch_2022_POCO} to reconstruct untextured meshes. Then we utilize our trained T.F. network to infer textures, by querying colors for the associated 3D point of each texture atlas pixel. In this way, both PointDreamer and the T.F. baseline adopt the same geometry module for a fair comparison. 
Please refer to our supplementary file for more details about the training and inference of our T.F. baseline.

\begin{table}[t]
  \centering
  \fontsize{9}{9}\selectfont
  \setlength{\tabcolsep}{1mm}
  
  {
    \caption{Comparisons on textured mesh reconstruction on ShapeNetCoreV2, GSO and Omniobject3D datasets. T.F. refers to Texture Field. Note that both T.F. and our PointDreamer adopt POCO~\cite{Boulch_2022_POCO} for geometry reconstruction for a fair comparison.
    }
    \label{tab:shapenet}%
    \begin{NiceTabular}{c|c|cccc}
        \toprule
    \textbf{ShapeNet Cat.} & \textbf{Method} & \textbf{PSNR ↑} & \textbf{SSIM ↑} & \textbf{LPIPS ↓} & \textbf{FID ↓} \\
    \midrule
    \multirow{4}[2]{*}{Chairs} & SPR   & 24.10 & 0.931 & 0.092 & 12.03 \\
          & NKSR  & 23.05 & 0.931 & 0.096 & 29.11 \\
          & T.F.  & 26.12 & 0.947 & 0.066 & 6.83 \\
          & Ours  & \cellcolor[rgb]{ .906,  .902,  .902} \textbf{26.29} & \cellcolor[rgb]{ .906,  .902,  .902} \textbf{0.952} & \cellcolor[rgb]{ .906,  .902,  .902} \textbf{0.057} & \cellcolor[rgb]{ .906,  .902,  .902} \textbf{4.90} \\
    \midrule
    \multirow{4}[2]{*}{Cars} & SPR   & \cellcolor[rgb]{ .906,  .902,  .902} \textbf{23.46} & 0.929 & 0.088 & 43.89 \\
          & NKSR  & 21.01 & 0.919 & 0.096 & 131.05 \\
          & T.F.  & 22.42 & 0.919 & 0.087 & 38.08 \\
          & Ours  & 22.78 & \cellcolor[rgb]{ .906,  .902,  .902} \textbf{0.930} & \cellcolor[rgb]{ .906,  .902,  .902} \textbf{0.073} & \cellcolor[rgb]{ .906,  .902,  .902} \textbf{12.41} \\
    \midrule
    \multirow{4}[2]{*}{Motorbikes} & SPR   & 15.09 & 0.809 & 0.214 & 106.09 \\
          & NKSR  & 18.13 & 0.905 & 0.101 & 175.42 \\
          & T.F.  & 21.17 & 0.926 & 0.064 & 40.06 \\
          & Ours  & \cellcolor[rgb]{ .906,  .902,  .902} \textbf{21.20} & \cellcolor[rgb]{ .906,  .902,  .902} \textbf{0.930} & \cellcolor[rgb]{ .906,  .902,  .902} \textbf{0.056} & \cellcolor[rgb]{ .906,  .902,  .902} \textbf{28.42} \\
    \midrule
    \midrule
    \textbf{Dataset} & \textbf{Method} & \textbf{PSNR ↑} & \textbf{SSIM ↑} & \textbf{LPIPS ↓} & \textbf{FID ↓} \\
    \midrule
    \multirow{4}[2]{*}{GSO} & SPR   & 27.11 & 0.907 & 0.120 & 27.43 \\
          & NKSR  & 24.65 & 0.900 & 0.118 & 36.51 \\
          & T.F.  & 25.53 & 0.894 & 0.128 & 46.44 \\
          & Ours  & \cellcolor[rgb]{ .906,  .902,  .902} \textbf{27.24} & \cellcolor[rgb]{ .906,  .902,  .902} \textbf{0.923} & \cellcolor[rgb]{ .906,  .902,  .902} \textbf{0.083} & \cellcolor[rgb]{ .906,  .902,  .902} \textbf{9.32} \\
    \midrule
    \multirow{4}[2]{*}{OmniObject3D} & SPR   & \cellcolor[rgb]{ .906,  .902,  .902} \textbf{30.20} & 0.927 & 0.100 & 40.13 \\
          & NKSR  & 26.77 & 0.919 & 0.105 & 50.14 \\
          & T.F.  & 28.08 & 0.914 & 0.112 & 65.61 \\
          & Ours  & 29.78 & \cellcolor[rgb]{ .906,  .902,  .902} \textbf{0.941} & \cellcolor[rgb]{ .906,  .902,  .902} \textbf{0.068} & \cellcolor[rgb]{ .906,  .902,  .902} \textbf{18.23} \\
    \bottomrule
    \end{NiceTabular}%
    }%

\end{table}%

\subsection{Main Results}

\subsubsection{Qualitative comparisons}
We present visual results in Figure~\ref{fig:results}. Since DHSP3D is much slower than other methods (4-6 hours per shape compared to $<$1 min), we randomly select two objects from GSO dataset for efficiency during comparison. \textbf{We highly recommend referring to our supplementary materials for more visual results and a 360° video.}

Figure~\ref{fig:results} (a)-(d) show that our method, though zero-shot, achieves much clearer and more realistic textures than baselines, 
thanks to the 2D diffusion prior. 
Besides, SPR and NKSR, which both rely on per-point normals as input, sometimes yield redundant geometry mainly due to the wrongly-estimated normal direction. The texture field baseline, which is trained on ShapeNetCoreV2, performs much worse on the newly-seen dataset of GSO and Omniobject3D, indicating a lack of generalization ability. %
In Figure~\ref{fig:results} (d), the red cup in the first row shows that DHSP3D does not support arbitrary topology. This is because it optimizes meshCNN and 2D XYZ map from the convex hull of the input point cloud, enforcing the output mesh to adhere to the hull's topology. Besides, DHSP3D suffers from less realistic textures with jagged or unclear edges, 
since its purely self-prior property only enables it to predict point colors considering corresponding neighbors, without the ability to \emph{dream} the unseen like our PointDreamer.

\subsubsection{Quantitative comparisons}
We summarize the quantitative results in Tables~\ref{tab:shapenet}, and \textbf{geometry evaluation results in our supplementary file}.
Our method outperforms baselines on most metrics and datasets, especially on the two perceptual metrics FID and LPIPS with a significant margin. However, regarding PSNR, which prioritizes pixel-level accuracy and thus may differ from human perception of quality, our method is sometimes outperformed by SPR. PSNR has been found to prefer blurry images~\cite{lpips}. We assume that SPR predicts a point's color by \emph{blending} nearby points' colors, leading to blurry textures but avoiding extreme pixel-level errors. On the contrary, our method \emph{dreams} the colors by its diffusion prior. While achieving visually better results, a few pixels with extremely high errors may lower the PSNR score.

\subsubsection{Multi-view Consistency}
\begin{figure}[t]
  \centering
    \includegraphics[width=\linewidth]{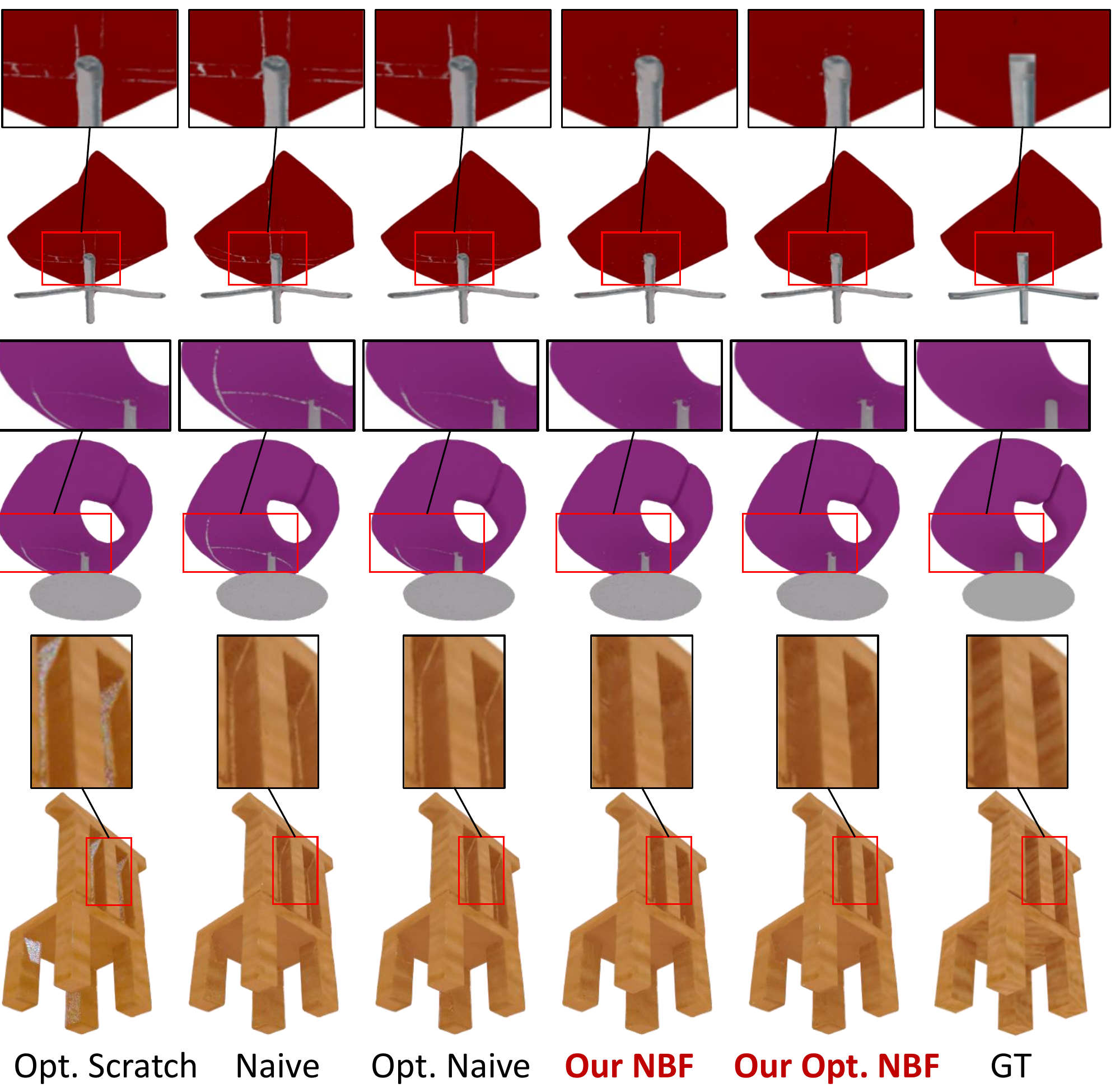}
  
  \caption{ Ablation comparisons between our proposed NBF strategy and other unprojection strategies. 
  }
  \label{fig:NBF_results}%
\end{figure}

Figure~\ref{fig:results} (e) and our supplementary video show that, our method achieves high multi-view consistency. The reason is two-fold: (1) Unlike image-conditioned 3D generation, it's easier to maintain multi-view consistency when reconstructing point clouds, since the input point cloud already determines the global shape and colors, leaving only local details to be \textit{dreamed} by the model. (2) For inconsistencies in local details, our proposed Non-Border-First unprojection effectively addresses this issue; see Figure~\ref{fig:NBF_results} (c) and Table~\ref{tab:unproject}. 

\subsubsection{\yq{More Visual Presentation}}
\begin{figure*}
    \centering
    \includegraphics[width=0.75\linewidth]{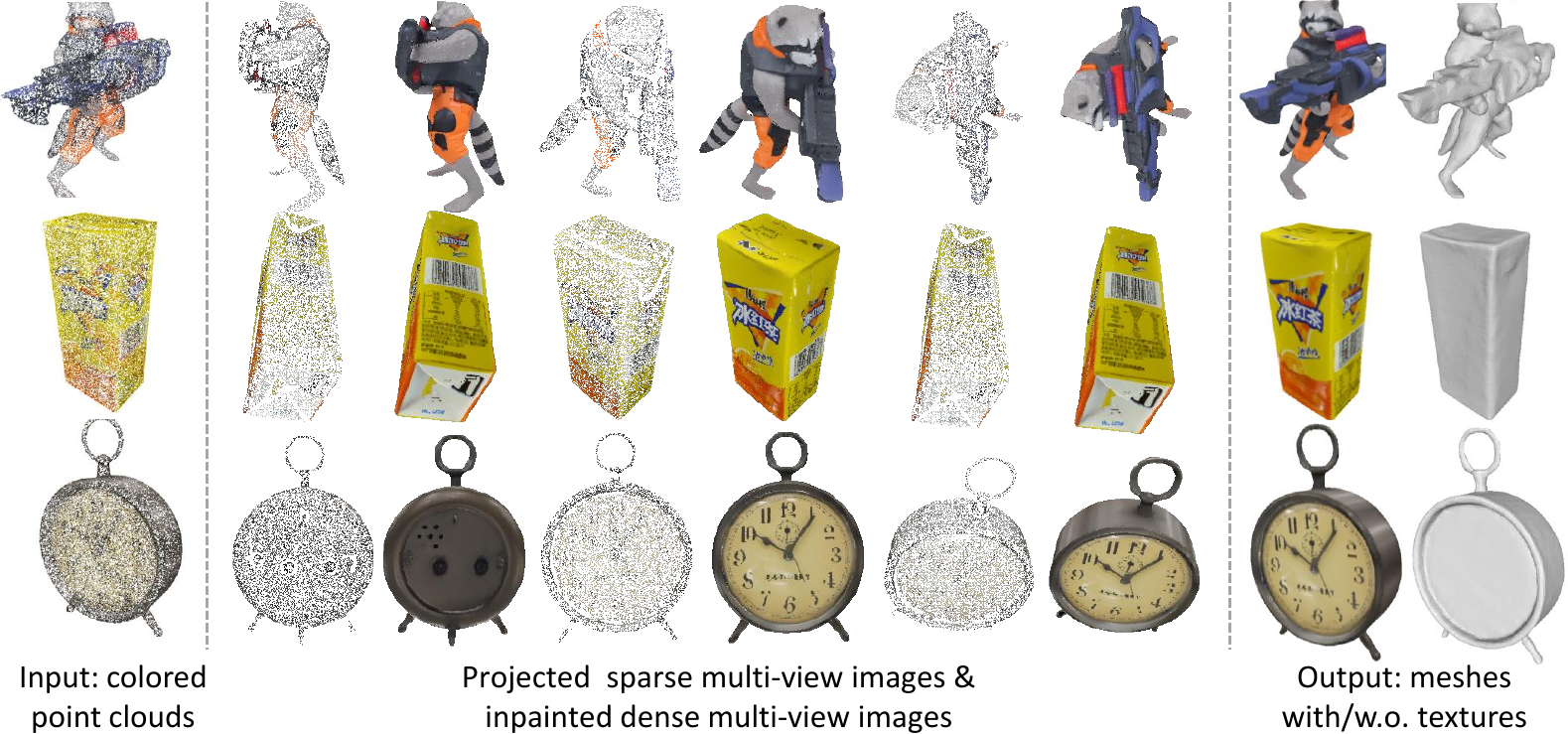}
    \caption{Visual presentation of the input point clouds, intermediate sparse and dense multiview images, and the output meshes.}
    \label{fig:intermediate}
\end{figure*}

\yq{For readers to better understand our pipeline, we present the input point cloud, intermediate multiview images, and output meshes in Figure~\ref{fig:intermediate}}.

\begin{table}[t]
    \fontsize{9}{9}\selectfont
  \setlength{\tabcolsep}{1mm}
  \centering
  \caption{Anti-noise comparisons on ShapeNetCoreV2 Chairs. `Noisy' means adding Gaussian noise with a standard deviation of 0.005 to the input.}
    \begin{NiceTabular}{cc|cccc}
 
    \toprule
    Input & Method & PSNR ↑ & SSIM ↑ & LPIPS ↓ & FID ↓ \\
    \midrule
    Noisy & SPR   & 19.64 & 0.884 & 0.170 & 65.25 \\
    Noisy & NKSR  & 22.79 & 0.929 & 0.102 & 44.02 \\
    Noisy & T.F.  & 26.01 & 0.944 & 0.071 & 8.21 \\
    Noisy & Ours  & \cellcolor[rgb]{ .906,  .902,  .902} \textbf{26.26} & \cellcolor[rgb]{ .906,  .902,  .902} \textbf{0.952} & \cellcolor[rgb]{ .906,  .902,  .902} \textbf{0.057} & \cellcolor[rgb]{ .906,  .902,  .902} \textbf{4.93} \\
    \midrule
    Clean & T.F.  & 26.12 & 0.947 & 0.066 & 6.83 \\
    \bottomrule
    \end{NiceTabular}%

  \label{tab:noise}%
\end{table}%

\begin{table}[t]
  \centering
  
    \fontsize{9}{9}\selectfont
  \setlength{\tabcolsep}{1mm}
  \caption{Performance comparison on ShapeNetCoreV2 Chairs with increased input point cloud sparsity.}
  {
    \begin{NiceTabular}{cc|cccc}
   
      \toprule
    Point Number & Method & PSNR ↑ & SSIM ↑ & LPIPS ↓ & FID ↓ \\
    \midrule
    30k   & Ours  & \cellcolor[rgb]{ .906,  .902,  .902} \textbf{26.29} & \cellcolor[rgb]{ .906,  .902,  .902} \textbf{0.9524} & \cellcolor[rgb]{ .906,  .902,  .902} \textbf{0.057} & \cellcolor[rgb]{ .906,  .902,  .902} \textbf{4.90} \\
    25k   & Ours  & 26.28 & 0.9517 & 0.058 & 5.24 \\
    20k   & Ours  & 26.26 & 0.9509 & 0.059 & 5.70 \\
    10k   & Ours  & 26.17 & 0.9477 & 0.064 & 7.79 \\
    \midrule
    30k   & T.F.  & 26.12 & 0.9467 & 0.066 & 6.83 \\
    \bottomrule
    \end{NiceTabular}%
    }

  \label{tab:sparsity}%
\end{table}%

\begin{table}[t]
  \centering

      \fontsize{9}{9}\selectfont
  \setlength{\tabcolsep}{1mm}
    \caption{
    Quantitative results of replacing our NBF unprojection strategy with others. ``Opt.'' denotes ``Optimize''. 
  }
    {
    \begin{NiceTabular}{c|cccc}
    \toprule
    Unprojection Module & PSNR ↑ & SSIM ↑ & LPIPS ↓ & FID ↓ \\
    \midrule
    Opt. Scratch & 26.19  & 0.9473  & 0.0592  & 5.26 \\
    Naive & 26.22  & 0.9500  & 0.0587  & 5.11 \\
    Opt. Naive & 26.25  & 0.9511  & 0.0580  & 4.94 \\
    \midrule
    NBF   & \cellcolor[rgb]{ .906,  .902,  .902} \textbf{26.27 } & 0.9512  & 0.0579  & 5.03 \\
    Opt. NBF  & 26.26  & \cellcolor[rgb]{ .906,  .902,  .902} \textbf{0.9516 } & \cellcolor[rgb]{ .906,  .902,  .902} \textbf{0.0574 } & \cellcolor[rgb]{ .906,  .902,  .902} \textbf{4.93} \\
    \bottomrule
    \end{NiceTabular}%
    }

  \label{tab:unproject}%
\end{table}%

\begin{figure}[t]
    \centering
\includegraphics[width=1.0\linewidth]{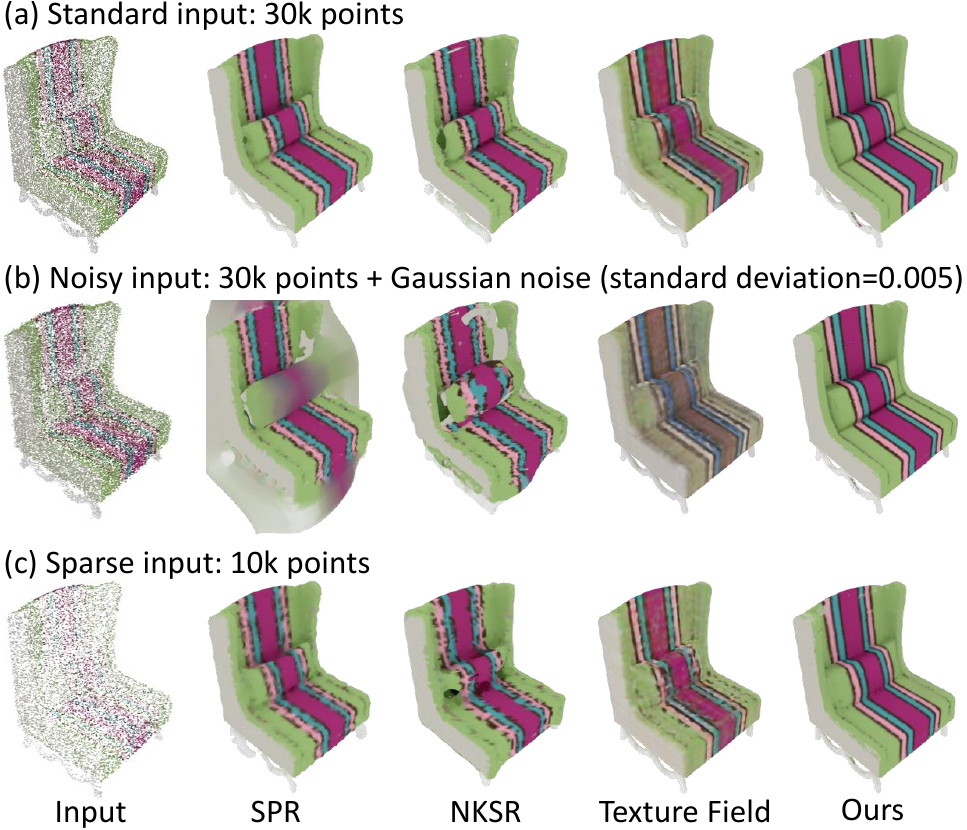}
    \caption{Our PointDreamer's robustness against noisy and sparse input. } 
    \label{fig:noisy_sparse}
\end{figure}

\begin{figure}[t]
    \centering
    \includegraphics[width=1.0\linewidth]{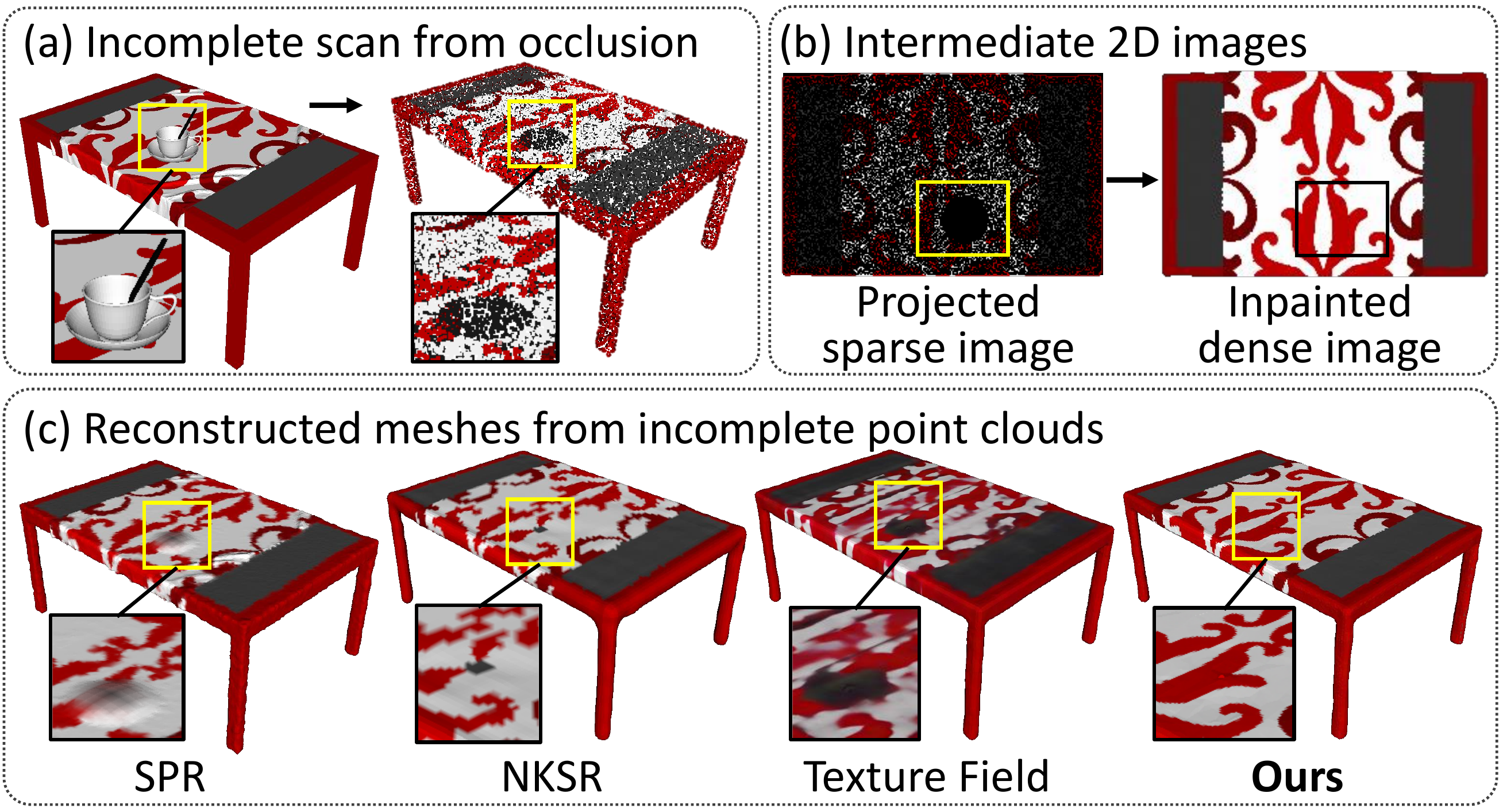}
    \caption{Our PointDreamer's unique completion ability. When the input point cloud is incomplete e.g. scanned from occluded objects as shown in (a), only our PointDreamer can plausibly complete the missing part with high-quality texture, as demonstrated in (c).  This is benefited from the strong inpainting power of the adopted 2D diffusion model, as illustrated in (b).}
    \label{fig:completion}
\end{figure}

\begin{figure}[t]
    \centering
\includegraphics[width=1.0\linewidth]{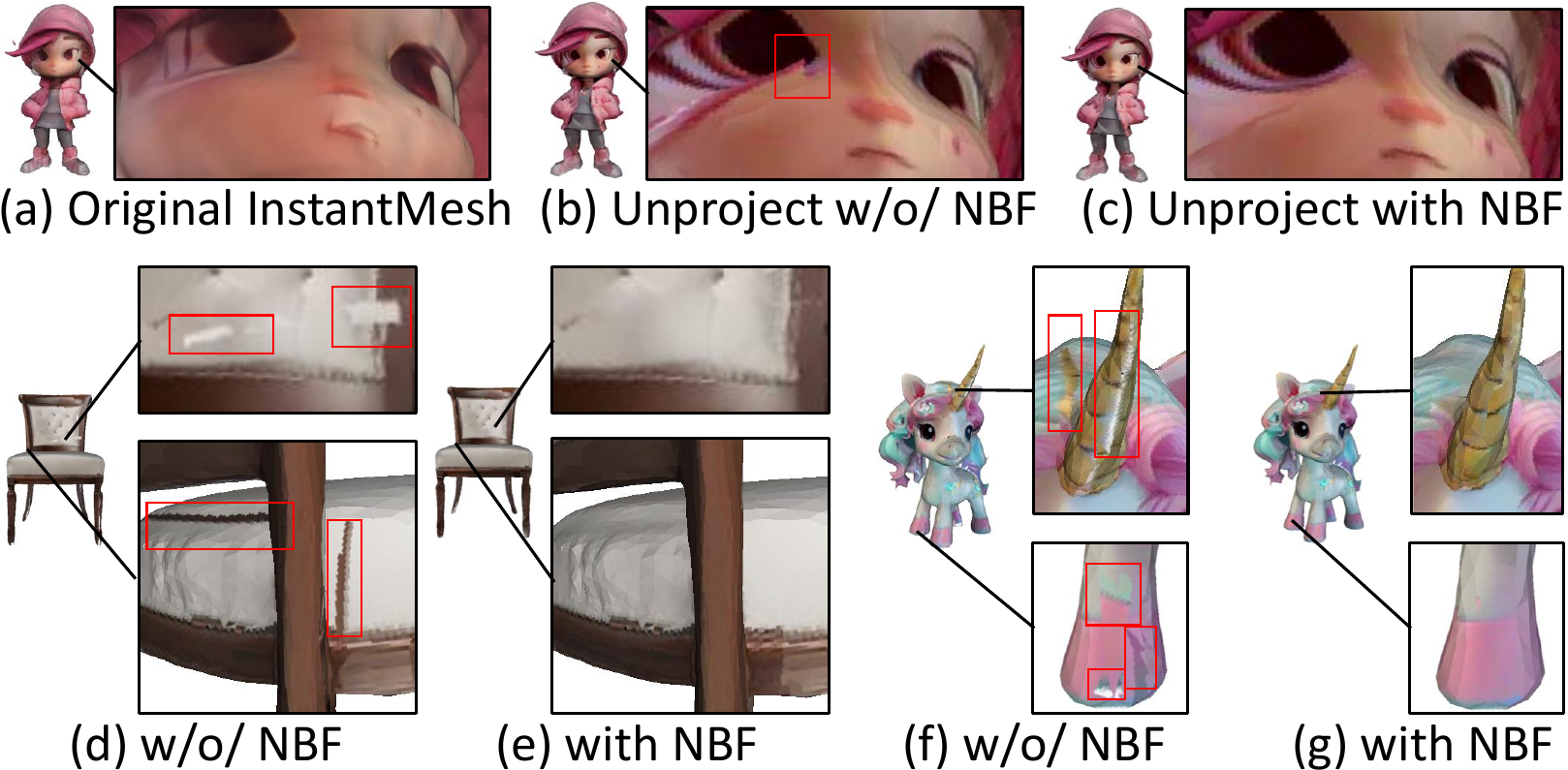}
    \caption{Examples of adopting our proposed NBF strategy on other methods (InstantMesh~\cite{xu2024instantmesh}). (a) The original mesh generated by InstantMesh looks blurry. (b) Directly unprojecting intermediate multiview images to the mesh alleviates the blurriness, but leads to artifacts at the border of the eye as marked in a red box. (c) Using our NBF unprojection strategy avoids such artifacts. (d-g) More examples. } 
    \label{fig:instantmesh_NBF}
\end{figure}

\subsection{Robustness against Degraded Input Quality}
\subsubsection{Anti-Noise Ability Analysis}
We compare 
our method and baseline methods by applying them on point clouds with manually added Gaussian noise (standard deviation = 0.005) from the chair category of ShapeNetCoreV2 dataset. 
Table~\ref{tab:noise} shows the results, where our method, even with noisy input, outperforms baselines with clean input. Figure~\ref{fig:noisy_sparse} (b) also shows our PointDreamer's strong anti-noise ability. While baseline methods exhibit significant performance drops given noisy input, our PointDreamer's performance degradation is minimal, even barely noticeable by human perception. 
\yq{We analyze that PointDreamer's relative robustness to noisy input stems from the strong diffusion prior. Although the input sparse images are noisy, the adopted 2D diffusion model is trained on a vast amount of clean and plausible 2D images with no noise. Consequently, the model yields relatively clean inpainted images as it learns to generate samples that follow its training data's distribution.}

\subsubsection{Sparsity Test} To evaluate our method's ability to deal with sparse input, we decrease the input point number from 30k to 10k gradually, and present the results in Table~\ref{tab:sparsity}. We can see only a small performance drop, and even with only 20k points as input, our method outperforms baseline methods with 30k points as input. When the input point number is decreased to 10k, our PointDreamer achieves higher PSNR and SSIM metrics compared to the Texture Field baseline with 30k points. We also provide visual comparisons in Figure~\ref{fig:noisy_sparse} (c), where our method significantly outperforms baseline methods with only 10k points as input. \yq{We analyze that the high generation capability of the 2D diffusion prior is the main reason for this robustness to sparse input data.}

\subsubsection{Texture Completion Ability}
Thanks to the inpainting power of the adopted 2D diffusion model, our PointDreamer has a unique ability to complete textures from incomplete scans. Figure~\ref{fig:completion} shows an example, where the point cloud of the table is incomplete since a small area is occluded by the cup during scanning. As shown in Figure~\ref{fig:completion} (a), the tablecloth is partially missing, revealing the black underside of the table below. 
For geometry reconstruction, such incompletion is less challenging: both our method and baseline methods manage to fill in this hole, as shown in Figure~\ref{fig:completion} (c). If facing more challenging cases,  we can also add a point cloud completion module~\cite{pcn} before mesh reconstruction.  %
However, texture completion is less explored, and baseline methods cannot effectively deal with it.  
Luckily, as shown in Figure~\ref{fig:completion} (b), our diffusion-based inpainting module addresses this issue. When projecting the incomplete 3D point cloud into a 2D sparse image from the top view, the occluded part is empty to be inpainted, thanks to our hidden point removal operation. The strong diffusion prior handles this empty area well and the inpainted dense image looks plausible. As a result, as shown in Figure~\ref{fig:completion} (c), our PointDreamer completes the missing area with high-quality texture compared to other methods.

\subsection{Ablation Study on NBF Unprojection}

To verify the effectiveness of our proposed NBF unprojection strategy, we conduct comparison experiments on the full test set of the chair category of ShapeNetCoreV2 dataset with Gaussian noise (standard deviation = 0.005).  
We present the comparison results in Figure~\ref{fig:NBF_results} and Table~\ref{tab:unproject}, including the following methods: 
\begin{itemize}
    \item \emph{Opt. Scratch}: Randomly initialize a texture atlas and then optimize it \yq{by minimizing the per-pixel MSE loss between the mesh renderings and the inpainted multiview images, as described in Section 3.4.2}; 
    \item \emph{Naive}: select the best views by direction priority;
    \item  \emph{NBF}: select the best views by our proposed Non-Border-First strategy;
    \item \emph{Opt. Naive}: \yq{Take \emph{Naive}'s obtained atlas as initialization and optimize it by minimizing the per-pixel MSE loss between the mesh renderings and the inpainted multiview images, as described in Section 3.4.2};
    \item \emph{Opt. NBF}: \yq{Take \emph{NBF}'s obtained atlas as initialization and optimize it by minimizing the per-pixel MSE loss between the mesh renderings and the inpainted multiview images' non-border areas.}
\end{itemize}

Figure~\ref{fig:NBF_results} shows that the naive unprojection yields the above-mentioned artifacts from border areas, which can be slightly reduced by further optimization. Only our NBF and Opt. NBF nearly eliminate such artifacts. Table~\ref{tab:unproject} also shows that our proposed NBF and Opt. NBF outperform other unprojection strategies.

\yq{Importantly, going beyond the specific colored-PC-to-mesh task, our proposed NBF unprojection strategy can effectively generalize to any method that textures an existing mesh given multiview images. In these methods, even with specifically designed techniques, the local border-area inconsistencies between multiview images and geometry can hardly be eliminated, which our NBF addresses.

We take InstantMesh~\cite{xu2024instantmesh} as an example. It reconstructs a textured 3D mesh from a single image, by first generating multiview images through a multiview diffusion model~\cite{shi2023zero123plus}, and then outputting the mesh via a large reconstruction model that directly predict the mesh's geometry and texture in 3D space. Note that the multiview diffusion technique is specifically designed for multiview-consistent generation.

We present visual comparisons in Figure~\ref{fig:instantmesh_NBF}. As mentioned in our introduction, learning colors in 3D space can result in blurry textures, as shown in Figure~\ref{fig:instantmesh_NBF} (a). When unprojecting the multiview images to the mesh using the naive unprojection strategy, the texture becomes sharper, as shown in Figure~\ref{fig:instantmesh_NBF} (b). However, as a cost, an artifact occurs in a border area, i.e. the border of the eye as marked in red. Our NBF strategy effectively solves this artifact, as shown in (c). More examples are provided in (d-g). These examples demonstrate that NBF can serve as a general solution to border-area artifacts in multiview-image-to-texture methods.}

\begin{figure}[t]
  \centering
    \includegraphics[width=\linewidth]{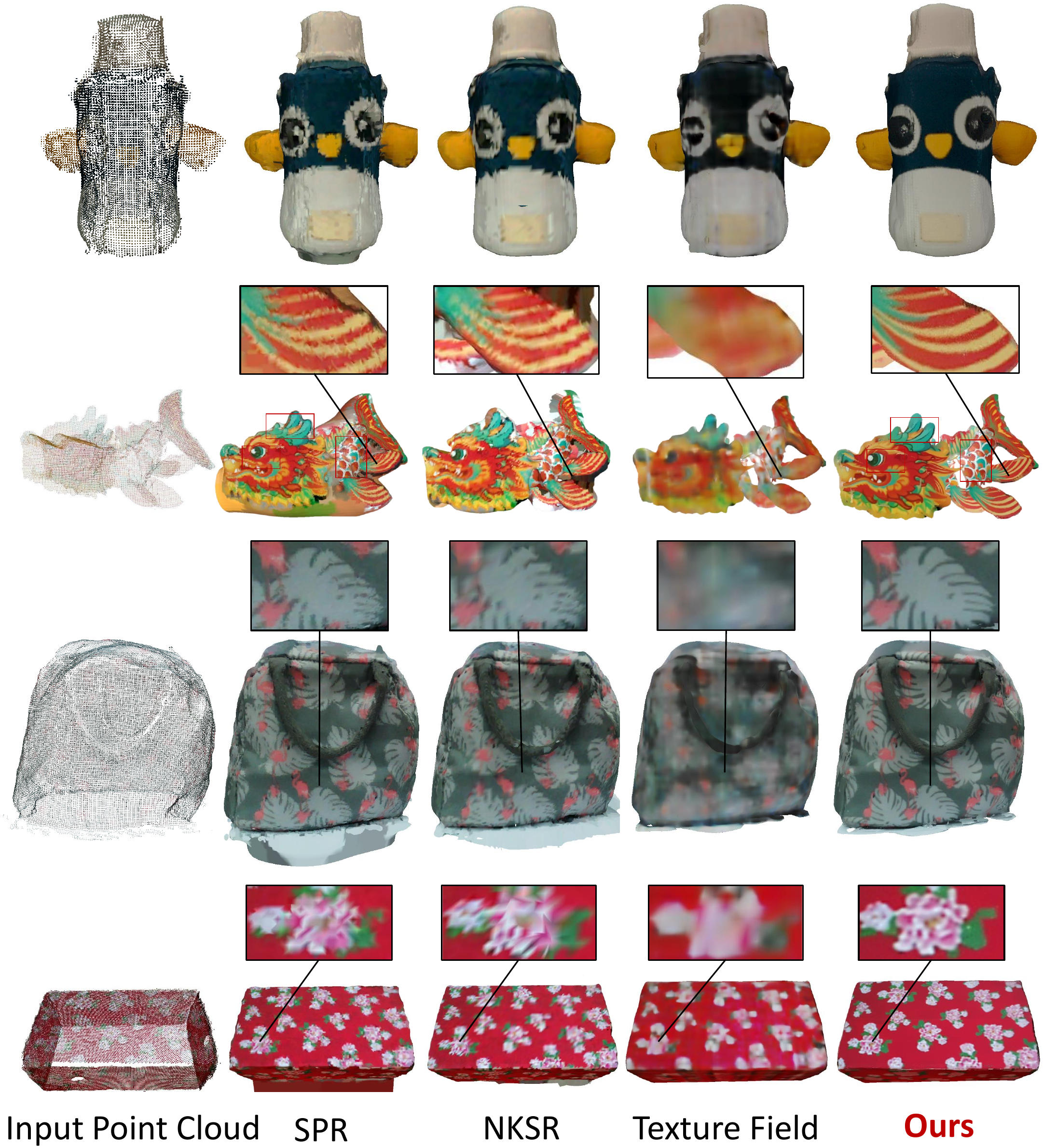}
  
  \caption{\yq{Comparison results with real-scanned point clouds as input. Better zoom in.}}
  \label{fig:real-scanned}%
\end{figure}

\yq{
\subsection{Real-life Data Experiments}
Above, we followed the common routine of point cloud reconstruction works~\cite{onet,Wei_2021_DHSP3D,Boulch_2022_POCO} to conduct experiments using point clouds sampled from existing meshes, because the calculation of existing quantitative evaluation metrics like PSNR require ground-truth meshes. 

To better evaluate the effectiveness of our PointDreamer, we further capture point cloud inputs by scanning real-life objects with an Intel RealSense L515 LiDAR sensor. 
To ensure a more comprehensive evaluation, we select objects with diverse shapes and scales. As shown in Figure~\ref{fig:real-scanned}, the bottle in the top row is approximately a cylinder, the ``Aoyu'' in the second row and the box in the bottom row mainly consist of planes, and the bag in the third row exhibits a less regular shape. The bottle is the smallest in real life, leading to the noisiest input point cloud due to the limited sensor resolution. Since no ground-truth meshes are available for quantitative assessment of these scanned objects, we present the qualitative results in Figure~\ref{fig:real-scanned}.

}

\yq{As shown in Figure~\ref{fig:real-scanned}, other methods suffer from different levels of blurriness or jagged effects. Compared to them, our PointDreamer archives clearer texture reconstruction, thanks to the diffusion prior. 
}

\section{Conclusion, Limitations and Future Work}
\para{Conclusion.}
We propose PointDreamer, a novel framework for textured mesh reconstruction from colored point cloud with SoTA performance. By utilizing diffusion-based 2D inpainting, it (1) reconstructs clear and high-quality mesh textures, addressing the common blurring issue; (2) shows high robustness against sparse, noisy, or even incomplete input, and (3) works in a zero-shot manner, requiring no extra training.  %
We also propose a novel ``Non-Border-First'' strategy to unproject the colors of predicted 2D images back to 3D space. This strategy is the first to address the border-area artifact issue, which is less-explored but commonly-occurred in methods that generate 3D textures from multiview images.

\para{Limitations and Future Work.} (1) Like all other methods producing meshes from multi-view images, our method cannot perfectly color the small unseen areas. (2) The adopted inpainting module DDNM, though with SoTA performance, cannot perfectly inpaint all cases, especially those with extremely complex, irregular patterns or very fine details. 
In the future, we plan to develop an adaptive camera placement strategy to minimize invisible regions and better cover the entire mesh. Additionally, we aim to further investigate a relighting-supportive reconstruction approach: disentangling illumination effects from the input points' colors, and reconstructing albedo colors and material information for the mesh, to better support relighting in graphics pipelines.

\bibliographystyle{IEEEtran}
\bibliography{main}

\clearpage
\appendices

\section{Geometry Evaluation Results}

To compare the reconstructed geometry quality of PointDreamer and baseline methods while ignoring textures, 
we report geometry evaluation results in Table~\ref{tab:geo_eval}. Same as our manuscript, we use ShapeNetCoreV2~\cite{shapenet2015} (Chair, Car, Motorbike categories), Google Scanned Objects~\cite{downs2022google} and Omniobject3d~\cite{wu2023omniobject3d} datasets. We follow POCO~\cite{Boulch_2022_POCO} to use commonly-used metrics of Chamfer L1-distance ×100 (CD), normal consistency (NC), %
and F-Score with threshold value 1\% (FS). See POCO for more details about these metrics. Note that, as mentioned in our manuscript, we use POCO as the geometry extraction module for both our PointDreamer and the Texture Field baseline.

\section{More Visual Results}

\begin{figure}[t]
  \centering
  \includegraphics[width=1.0\linewidth]{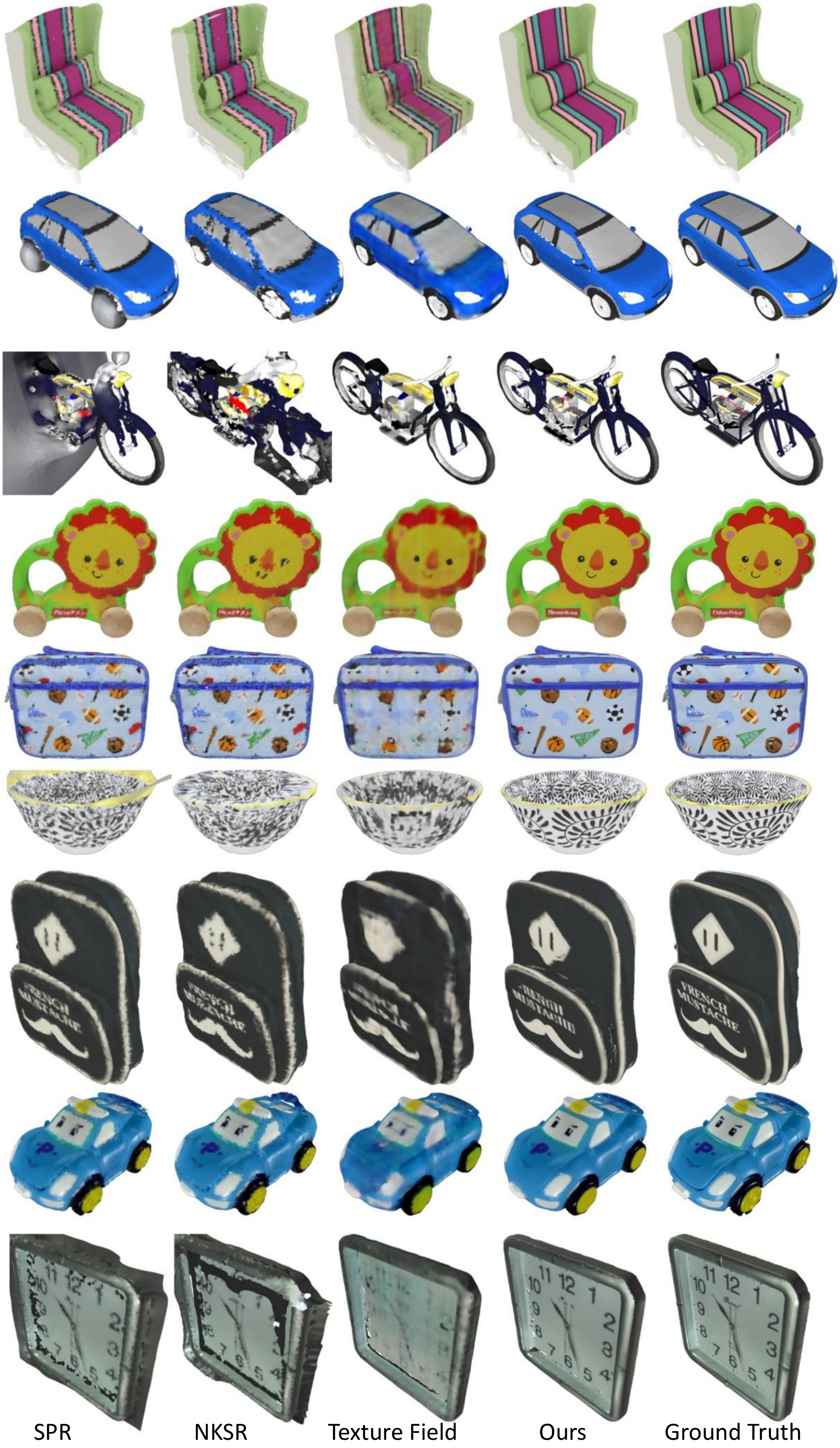}
  \caption{More visual comparisons with baseline methods.  Row 1-3: ShapeNetV2 dataset. Row4-6: GSO dataset. Row 7-9: OmniObject3d dataset. }
  \label{fig:more_results}%
\end{figure}

Figure~\ref{fig:more_results} shows more visual comparisons of our PointDreamer against baseline methods.

\section{\yq{Method Component Analysis}}
\yq{
We here compare different implementations for our geometry extraction and inpainting sub-modules by replacing one module at a time while keeping everything else unchanged. All experiments in this section are conducted on the full test set of the chair category of ShapeNetCoreV2 dataset with Gaussian noise (standard deviation = 0.005).

\subsection{Geometry Extraction Module Replacement Experiment}
To investigate how the geometry extraction module influences the final reconstructed textured mesh, we explore three different geometry extraction methods: POCO~\cite{Boulch_2022_POCO} (as adopted in our manuscript), SPR~\cite{kazhdan2013screened} and depth inpainting.
}
\subsubsection{Depth-Inpainting for geometry extraction}
\begin{figure}[t]
  \centering
    \includegraphics[width=\linewidth]{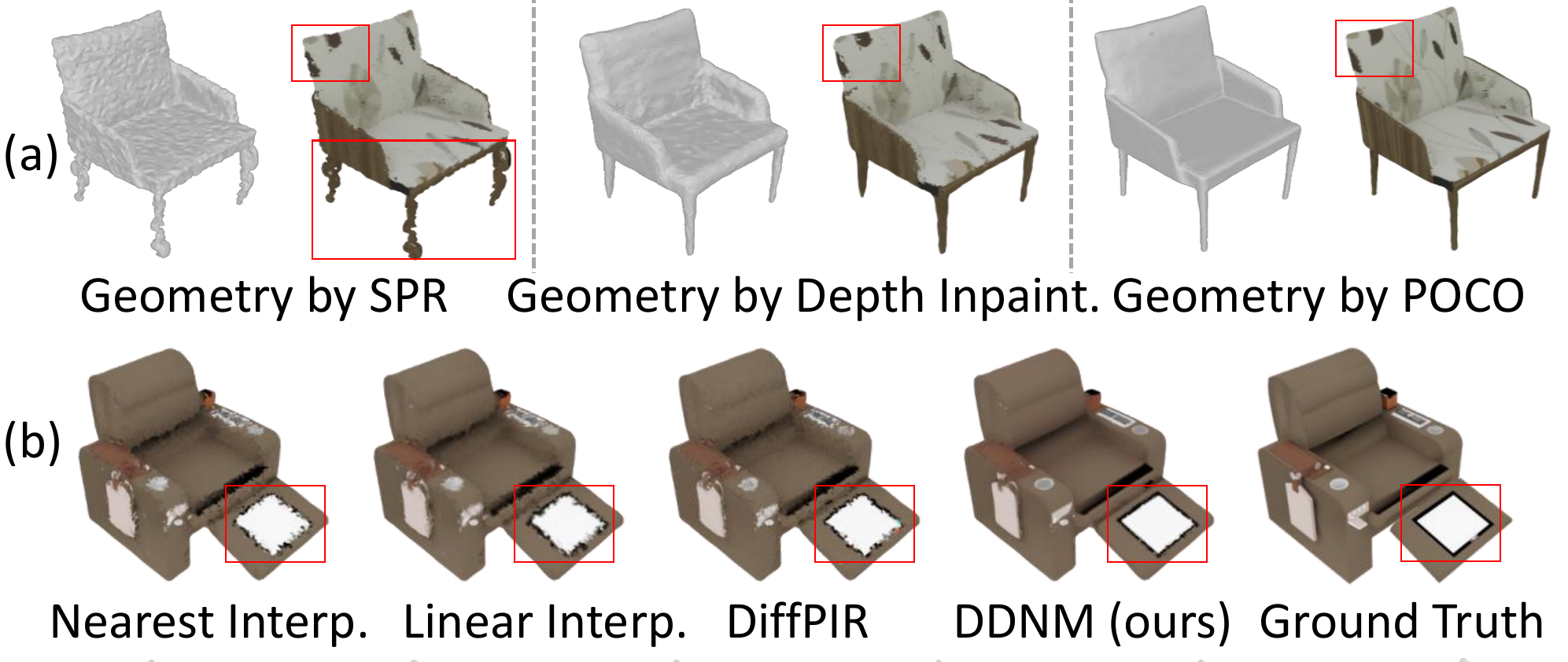}
  \vspace{-2em}
  \caption{ \yq{Sub-Module Replacement Results: (a) Geometry module. (b)
Inpainting module. }
  }
  \label{fig:submodule}%
\end{figure}

\begin{figure*}[t]
  \centering
  \includegraphics[width=0.9\linewidth]{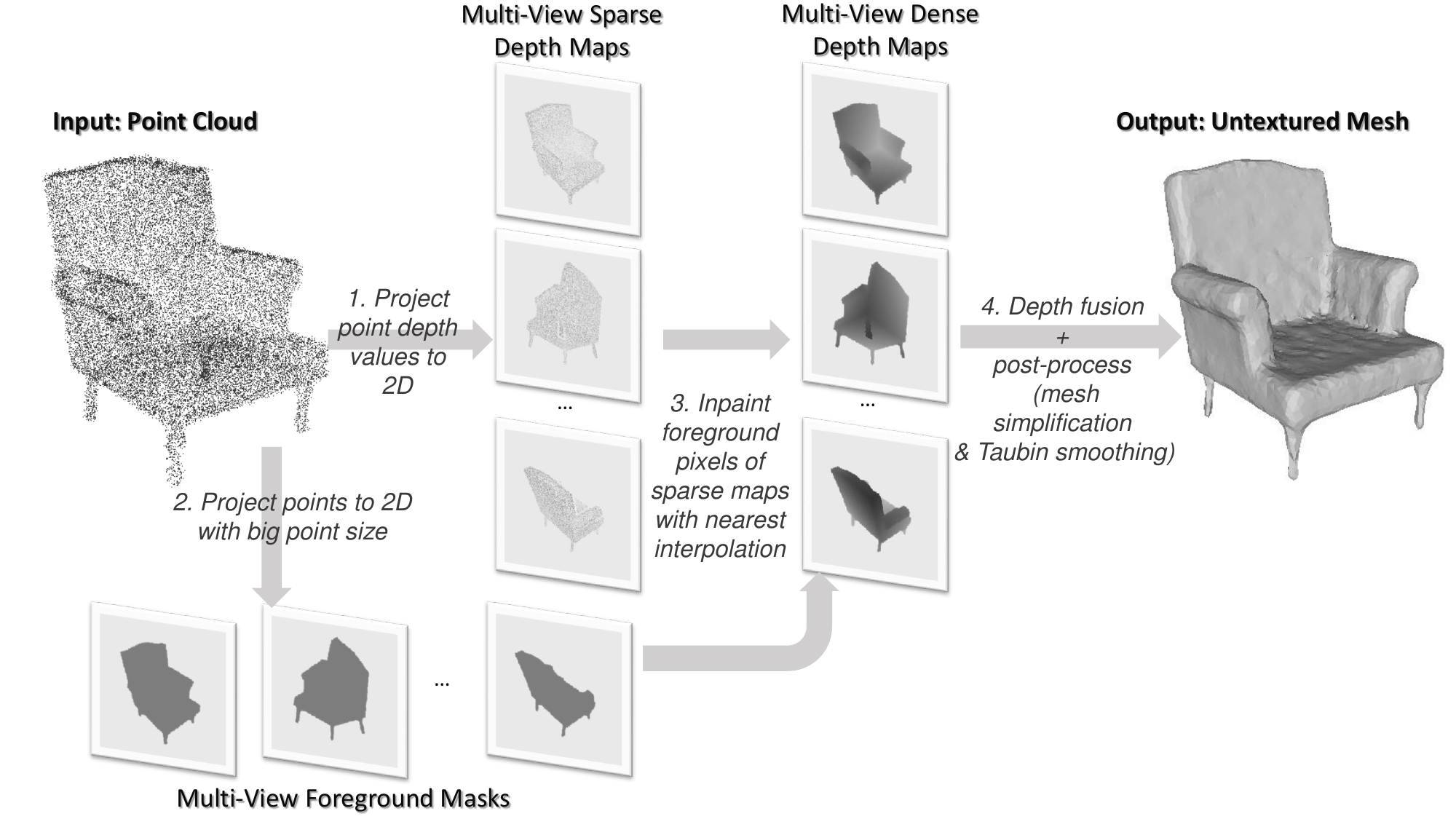}
  \caption{Pipeline of extracting geometry by depth inpainting. }
  \label{fig:depth_inpainting}%
\end{figure*}

With 2D inpainting adopted as the key for our texture reconstruction, we can also introduce it to geometry reconstruction, by inpainting 2D depth maps instead of RGB images.  Regarding this, we conduct experiments of depth inpainting, which refers to inpainting projected sparse 2D depth maps instead of RGB images, and then reconstructing geometry from the inpainted dense depth maps by depth fusion~\cite{pyfusion}. Specifically, we follow the following steps to reconstruct untextured meshes from input point clouds, see Figure~\ref{fig:depth_inpainting}:

\begin{enumerate}
  
\item We generate multi-view sparse depth maps by projecting 3D points to 2D, and assigning the value of each pixel as the depth value of the corresponding 3D point. Note that, similar to generating our sparse RGB images, hidden point removal is conducted for each viewpoint before projecting.

\item Since depth fusion requires depth maps' background pixels to have infinite values to produce a reasonable mesh, we generate foreground masks by projecting 3D points to 2D space with a relatively big point size, i.e. the number of 2D pixels occupied by each 3D point. In this way, most foreground pixels (pixels that should correspond to a point on the 3D mesh) can be occupied, and we use the closing operation of morphology to fill the rest small holes. In addition, since we use a big point size to generate the foreground mask, the mask would be bigger than the ground truth, so we shrink the generated mask by erosion.

\item We inpaint the foreground pixels of the sparse depth maps into dense ones by nearest interpolation considering efficiency.

\item We produce an untextured mesh by depth fusion based on the inpainted dense depth maps, and conduct mesh simplification~\cite{QEM} and Taubin Smooth~\cite{taubin} to it as post-processing, to get the final untextured mesh.

\end{enumerate}
\yq{
\subsubsection{Results}

We present the visual comparisons in Figure~\ref{fig:submodule} (a) and quantitative results in Table~\ref{tab:geo}, respectively. %
As can be seen, POCO, as a state-of-the-art deep-learning-based surface reconstruction approach, outperforms the other two methods by a significant margin, especially regarding FID. Both SPR and Depth Inpainting suffer from noisy geometry and thus low-quality textures.
This indicates that a higher-performance geometry extraction module contributes to a more refined reconstructed textured mesh.

\subsection{{2D Inpainting Module Replacement Experiment}}
We compare different inpainting modules in our pipeline, including nearest interpolation, linear interpolation,  DiffPIR~\cite{DiffPIR}, and our adopted DDNM, where DiffPIR is another diffusion-based 2D image restoration method with inpainting ability. 
We provide the visual comparisons in Figure~\ref{fig:submodule} (b) and quantitative results in Table~\ref{tab:inpaint}.  As can be seen, the other three methods, though perform overall reasonably, fail to produce as clear textures as DDNM, leading to a lower quantitative score. This indicates the importance of a strong inpainting module to the reconstruction performance. 
}

\section{More Experimental Results}
\subsection{Effect of Different K Values (Number of Viewpoints for Projection and Inpainting)}
\begin{figure*}[t]
  \centering
\includegraphics[width=0.6\linewidth]{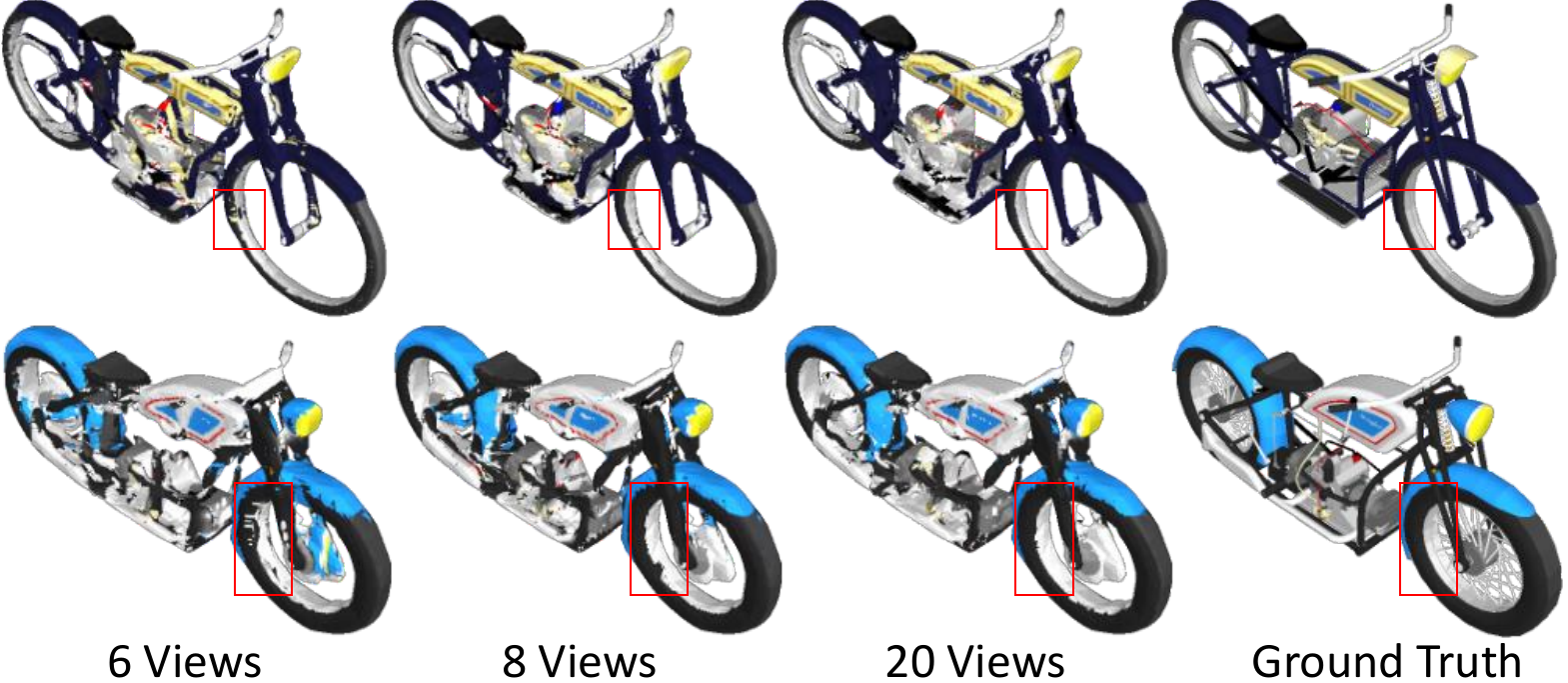}
  \caption{Visual comparisons of our reconstructed meshes with different $K$ values (numbers of viewpoints for projection and inpainting). An insufficient number of views would lead to artifacts in invisible or occluded areas.%
  }
  \label{fig:view_num}%
\end{figure*}

To investigate the effect of different numbers of viewpoints (denoted as $K$) for projection and inpainting, we conduct experiments on the motorbike category or ShapeNetCoreV2 dataset by using different $K$ values including 6, 8, and 20. Table~\ref{tab:view_num} shows the distribution of cameras for each setting, together with the quantitative results. We can see that more views contribute to a slightly higher reconstruction quality. Visual comparisons in Figure~\ref{fig:view_num} also show that, an insufficient number of views would produce artifacts in invisible or occluded areas, thus impacting the performance.

Considering that using $K=20$ views is only slightly better than setting $K=8$, but inpainting more views' images can be much more time-consuming, we set the number of views to be 8 for most datasets in our manuscript to balance both effectiveness and efficiency. The only exception is the motorbike category from the ShapeNetCoreV2 dataset, for which we use the 20 views instead, considering motorbikes' more complex geometry and topology.

\subsection{Impact of Degraded Input Quality: More Visual Comparisons}
\subsubsection{{Anti-Noise Ability Analysis}}
\begin{figure}[t]
  \centering
    \includegraphics[width=1.0\linewidth]{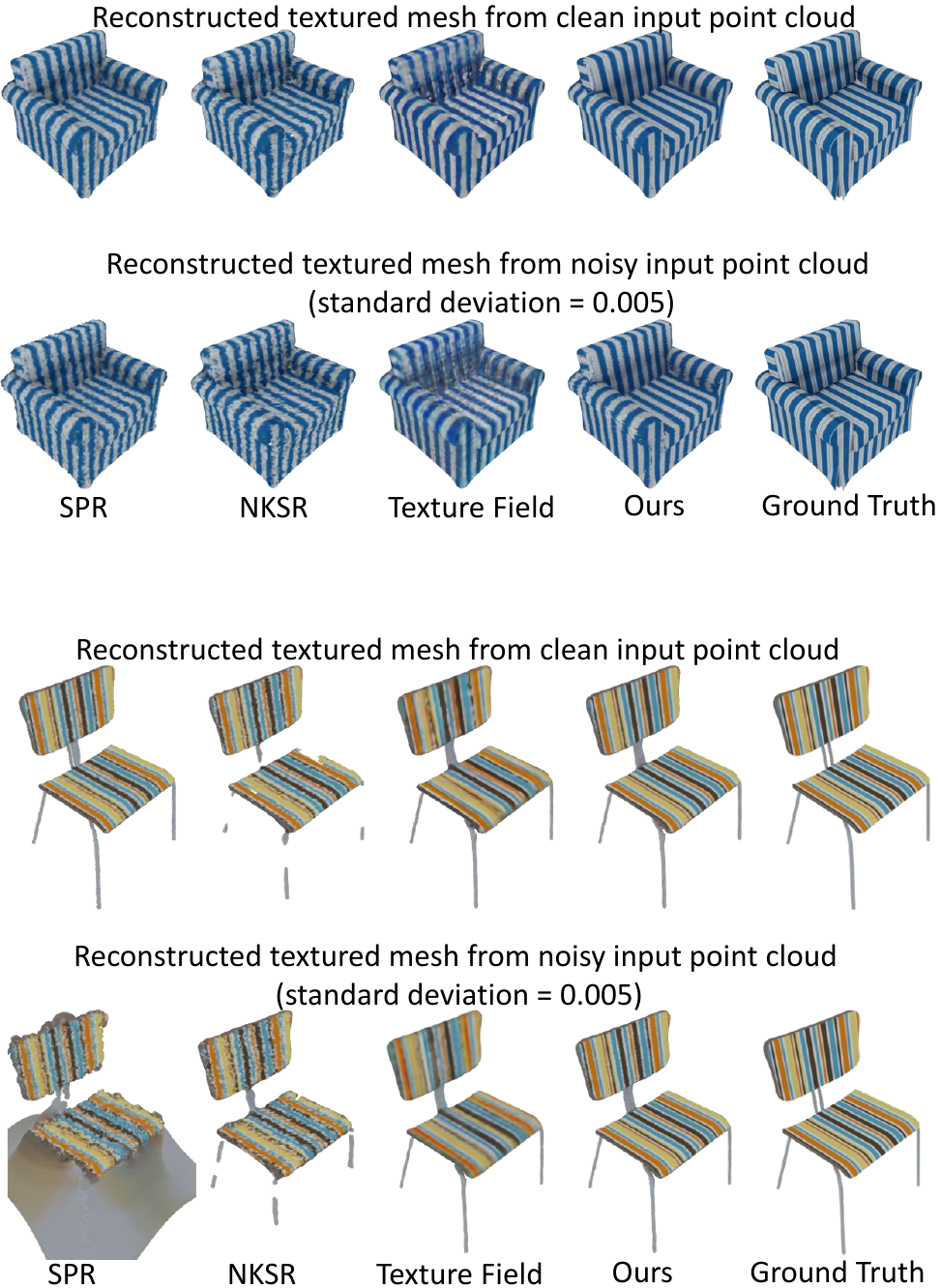}
  \caption{Visual comparisons of our PointDreamer's and baseline methods' reconstructed meshes, with noisy or clean point clouds as input. Our PointDreamer shows a strong anti-noise ability by producing high-quality textures even with noisy input. }

  \label{fig:noisy}%
\end{figure}

Figure~\ref{fig:noisy} presents the visual comparisons of our method and baseline methods' reconstructed meshes from noisy or clean input point clouds. Note that the reconstructed untextured meshes of noisy and clean inputs are the same, this is because our adopted POCO was trained with noisy input, so before reconstructing geometry, we manually add noise to the clean input. Overall, we have the following observations, which are consistent with the quantitative results (Table 2) in our manuscript:
\begin{enumerate}
  
\item As expected, all methods produce higher-quality textured meshes with clean input point clouds compared to noisy ones.

\item Our PointDreamer shows a relatively high anti-noise ability, where only a small performance drop is observed when giving noisy input.

\item Our PointDreamer with noisy input point clouds shows an even better visual effect compared to baseline methods with clean inputs.

\end{enumerate}

\subsubsection{{Sparsity Test}}
\begin{figure}[ht]
  \centering
  \includegraphics[width=0.8\linewidth]{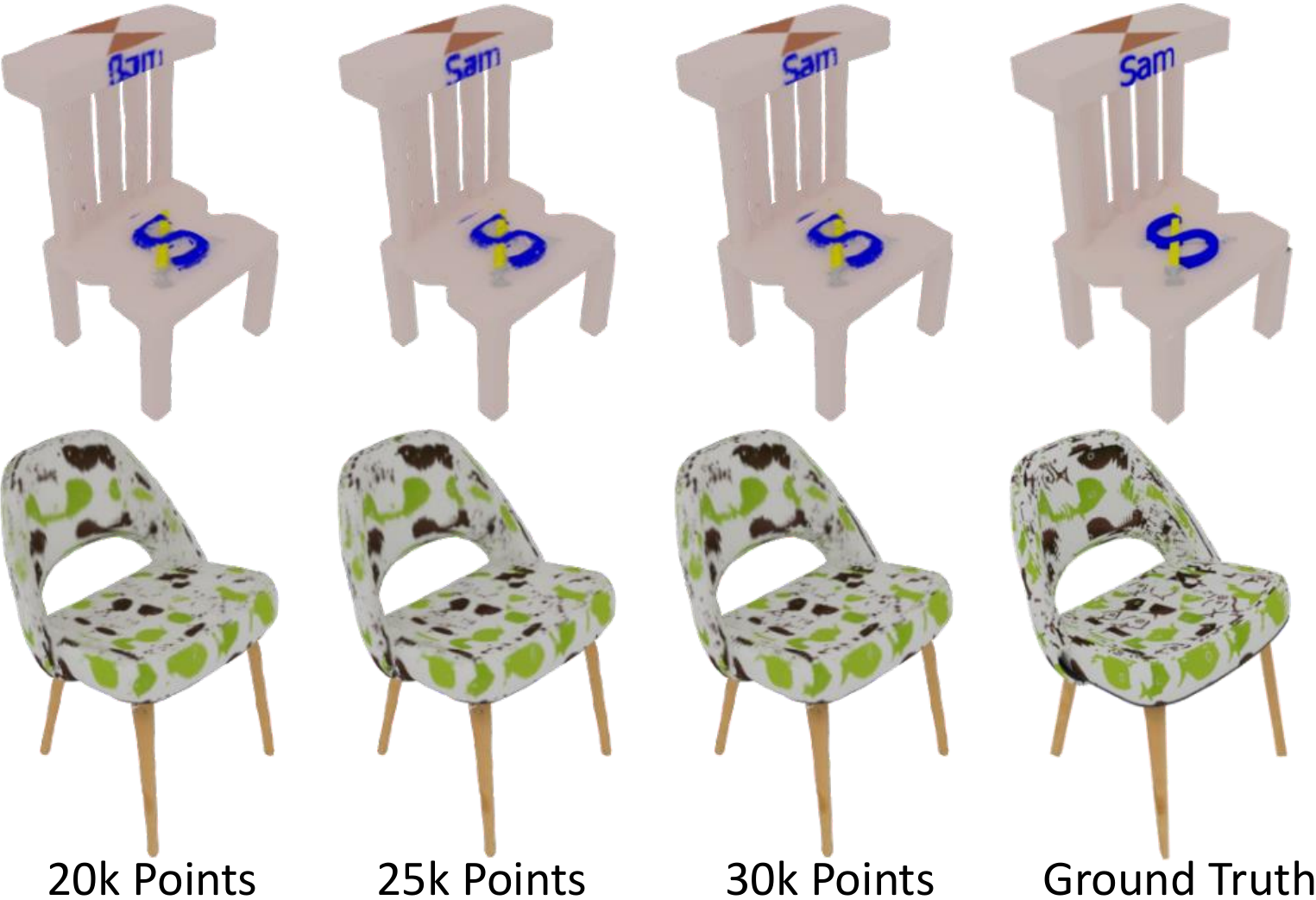}
  \caption{Visual comparisons of our PointDreamer's reconstructed meshes with different numbers of points as input. There is a small performance drop when adopting sparser input, which sometimes can be hard to notice by human eyes.}
  \label{fig:sparsity}%
\end{figure}

Figure~\ref{fig:sparsity} shows the visual comparisons of our PointDreamer's reconstructed meshes with different numbers of points as input. There is a very small performance drop introduced by decreasing the input point number, which can sometimes be hard to notice by human eyes. This indicates a relatively high robustness of our method towards varying degrees of input sparsity.

\subsection{\yq{Comparison with recent image-to-3D, text-to-3D, and mesh texturing methods}}

\begin{figure*}
    \centering
    \includegraphics[width=1\linewidth]{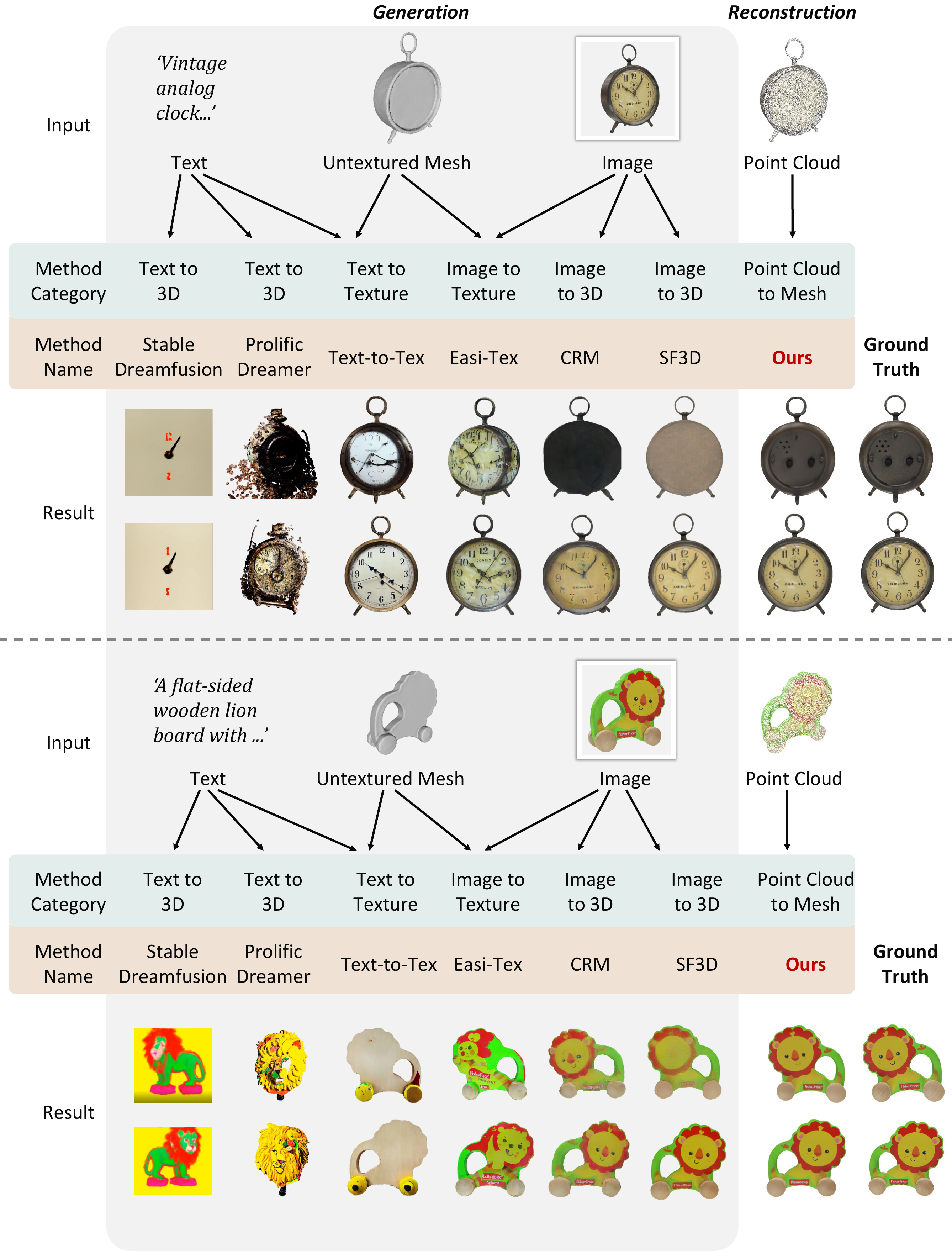}
    \caption{\yq{Comparison results with recent mesh texturing, image-to-3D, and text-to-3D methods.}}
    \label{fig:generation}
\end{figure*}

\yq{
\subsubsection{Experimental Setting}  
We select two of the most updated and representative methods from each relevant category (mesh texturing, image-to-3D, and text-to-3D), which results in the following comparison baselines: 
\begin{itemize}
    \item \textbf{StableDreamFusion}: A text-to-3D method based on SDS optimization. As the original  DreamFusion is not open-source, we utilized \href{https://github.com/ashawkey/StableDreamFusion}{an unofficial open-source implementation} with substantial community support (8.6k GitHub stars).
    \item \textbf{ProlificDreamer}: Another text-to-3D method leveraging SDS and further DMTet optimization.
    \item \textbf{Text-to-tex}: A text-conditioned mesh texturing (text-to-texture) method that employs diffusion-based inpainting.
    \item \textbf{Easi-tex}: An image-conditioned mesh texturing (image-to-texture) method, also utilizing diffusion-based inpainting.
    \item \textbf{CRM}: An image-to-3D method built upon multiview diffusion and a large feed-forward reconstruction network.
    \item \textbf{StableFast3D (SF3D)}: Another image-to-3D approach, characterized utilizing large feed-forward reconstruction network.
\end{itemize}

The above methods involve three kinds of inputs: 
\begin{itemize}%
    \item \textbf{Reference Image}: For methods requiring a reference image, we rendered images from the ground-truth mesh of the object.

    \item \textbf{Untextured Mesh}: For mesh texturing methods, we directly utilized the meshes reconstructed by our PointDreamer as the untextured mesh input.
        
    \item \textbf{Text Prompt}: To ensure fair comparison for text-based methods, we initially generated prompts by feeding the rendered ground-truth images into a large language model (Qwen2.5-VL-32B-Instruct). These automatically generated prompts were then manually refined to provide more accurate and detailed descriptions of the objects. The complete text prompts are provided below:
    \begin{itemize}
        \item Vintage analog clock: dark metal frame, matte khaki face with Arabic numerals. Hands at around 10:06:25. Small inset dial between center \& numeral 12; `CROSLEY' between center \& numeral 6. Base with two legs, top ring for hanging.

        \item A flat-sided wooden lion board with dual wooden wheels below. Features: yellow body, red petal mane, black eyes/orange nose/pink ears/blush marks. Green background \& handle on lion's back, 'Fisher-Price' in white on red at base under the lion.
    \end{itemize}

\end{itemize}

Note that most of these methods are relatively time-consuming (from about 15 minutes to more than an hour, compared to our method's 100 seconds).

\subsubsection{Experimental Results.} We present the comparison results in Figure~\ref{fig:generation}, which shows:
\begin{itemize}
    \item The text-to-3D methods StableDreamFusion and ProlificDreamer generate significantly different results compared to the ground truth, since they only rely on text for generation without any visual information. Even when not considering fidelity, the generation results suffer from severe implausibility (e.g. ProlificDreamer's generated clock) or even missing geometry (e.g. StableDreamFusion's generated clock) 
    \item Text-to-tex is free from the geometry unplausibility issue with a mesh provided as input, but still it can hardly faithfully reconstruct the texture.
    \item Easi-tex, though with an input image for reference, still struggles to generate high-fidelity textures, especially the backside without input information.
    \item CRM generates higher-fidelity results then previous methods, but it still can not faithfully reconstruct the back view. In addition, its generated textures are blurry.
    \item SF3D generates clearer textures from the front view, but its generated back views look highly implausible.
    \item Our PointDreamer can reconstruct the objects with the highest fidelity, in other words, most similar to the ground truth object.
\end{itemize}

In summary, only our PointDreamer can achieve relatively high fidelity reconstruction, both theoretically and experimentally.
}

\section{Implementation Details of Texture Field Baseline}
\subsection{{Network architecture}}
Inspired by works~\cite{gupta20233dgen,shen2021dmtet} that represent 3D information by a feature tri-plane, we adopt the network architecture of the open-source Convolutional Occupancy Network~\cite{convonet}, which also follows a tri-plane representation. Specifically, we modify its encoder from taking three-dimensional input (xyz) to six-dimensional input (xyzrgb). Then, we modify its one-dimensional output head for occupancy prediction to three-dimensional for RGB color prediction. %

\subsection{{Training}}
\begin{figure}[ht]
  \centering
  \includegraphics[width=\linewidth]{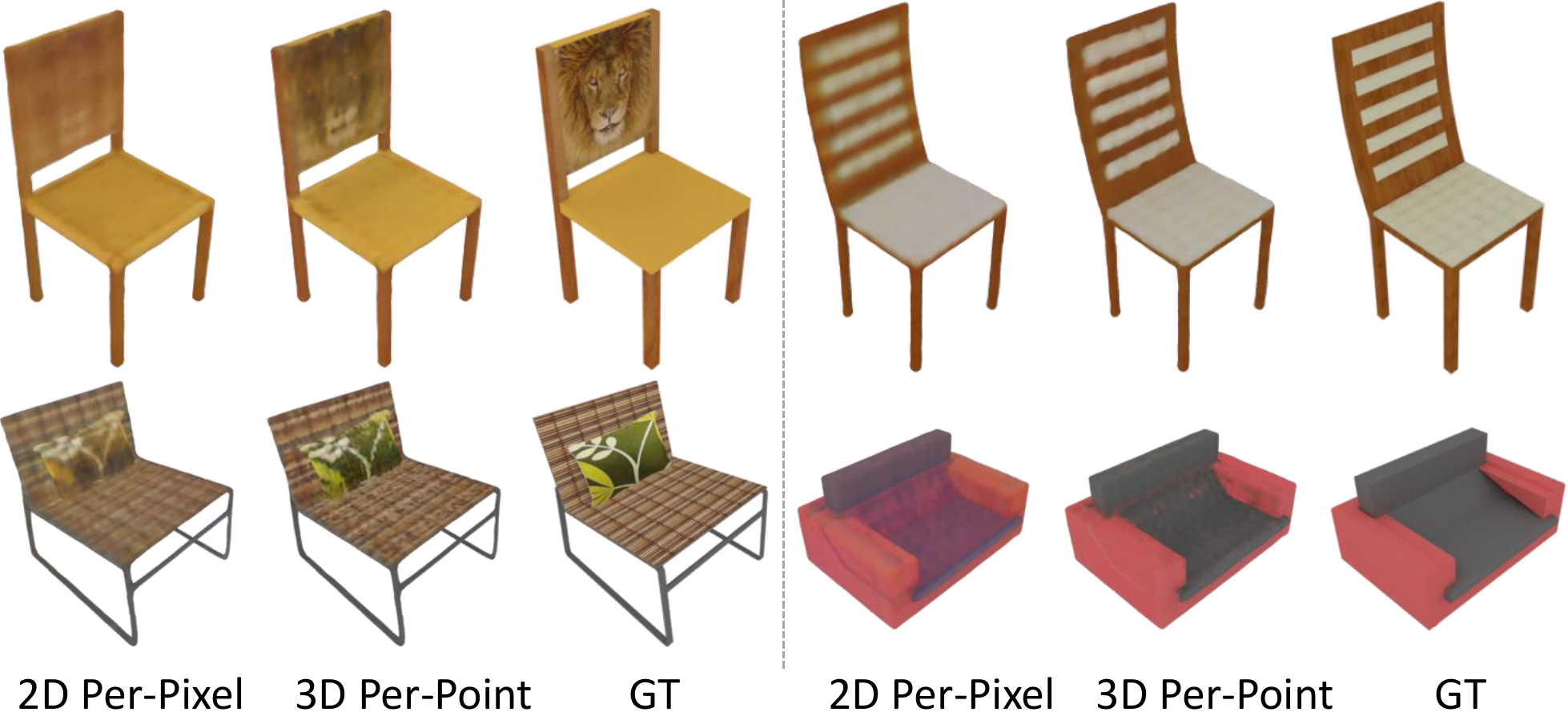}
  \caption{Visual comparisons of the Texture Field baseline trained with different losses. ``2D per-pixel'' denotes rendering the generated textured mesh to multi-view 2D images, and then calculating the MSE loss between the rendered and GT images. ``3D per-point'' denotes calculating MSE loss between the predicted and GT colors of sampled 3D points.
  }
  \label{fig:tf_loss}%
\end{figure}

We follow 3DGen~\cite{gupta20233dgen} to train the network by the per-3D-point MSE loss on predicted and GT colors of sampled 3D points. Additionally, we also tried to employ the per-2D-pixel MSE loss of 2D images rendered from GT meshes and predicted meshes by differentiable rendering, but the experimental results are worse, as shown in Figure~\ref{fig:tf_loss}. Therefore, we adopt the per-3D-point MSE loss in our manuscript. 
We train the texture field network on a single NVIDIA GeForce GTX 3090 GPU with a batch size of 24 and a learning rate of 0.0001 for 2592, 024 iterations (about 356 epochs).

\subsection{{Inference.}}
During inference, we first use POCO~\cite{Boulch_2022_POCO} to predict an untextured mesh from the input point cloud, and apply UV mapping to it by Xatlas~\cite{young2022xatlas}, which produces the 3D positions of each valid pixel in the texture atlas. We query the color of each of these 3D positions by our trained texture field network, to inpaint the texture atlas, so as to obtain the final textured mesh.

\section{Applications for large scenes}
\begin{figure}[h]
  \centering
  \includegraphics[width=\linewidth]{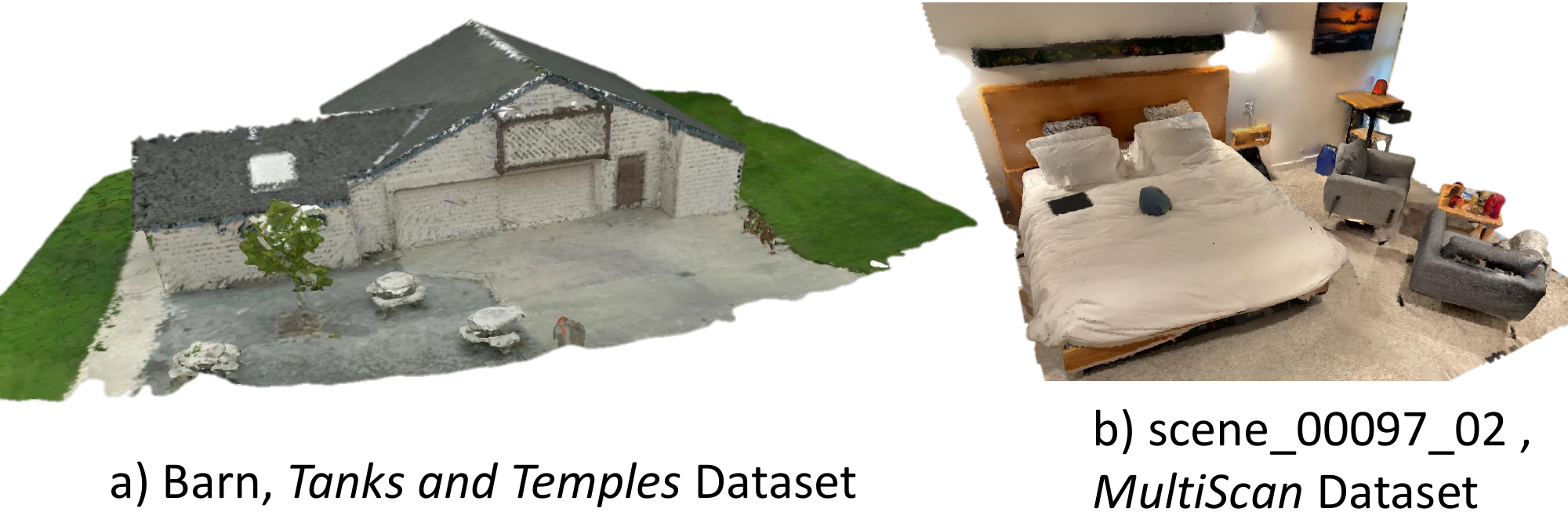}
  \caption{Our reconstruction results on scenes.}
  \label{fig:scene}
\end{figure}

PointDreamer, as a zero-shot method, can be adapted to indoor or outdoor scenes.  Figure~\ref{fig:scene} shows two examples from Tanks and Temples Dataset~\cite{tanks} and MultiScan~\cite{mao2022multiscan} dataset. Since our adopted diffusion model only supports images at $256 \times 256$ resolution, we downsample the scene point cloud before reconstruction. Also, our adopted POCO is relatively slow, especially with large-scale input point clouds. In the future, we may seek to develop a more large-scale-friendly version of PointDreamer, by adopting higher-resolution diffusion models and geometry reconstruction models designed for big scenes. 2D inpainting will remain our key idea, with only submodules replaced.

\section{\yq{Results on Objaverse Dataset}}
\begin{figure}[h]
  \centering
  \includegraphics[width=\linewidth]{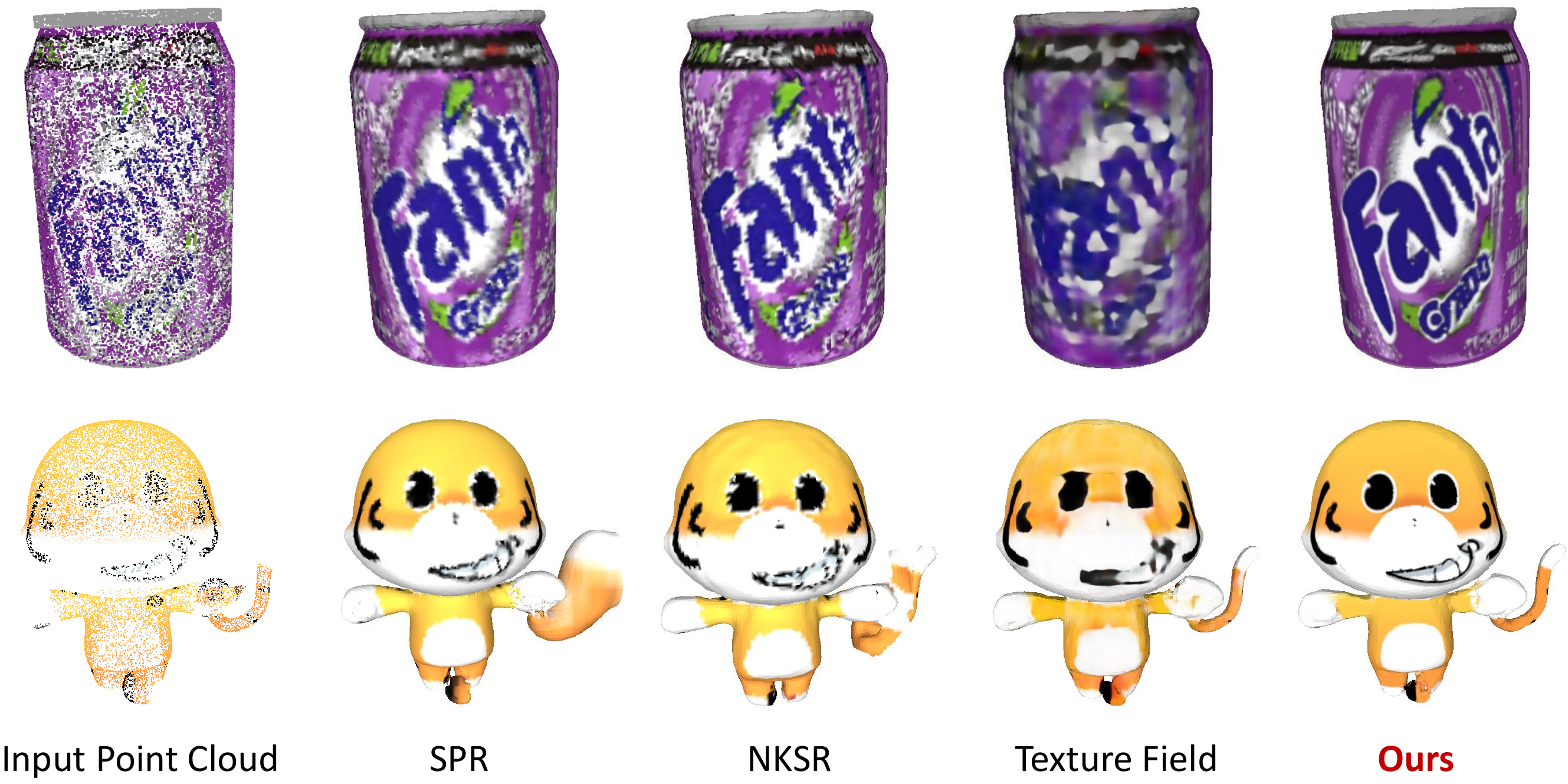}
  \caption{\yq{Results on the Objaverse dataset.}}
  \label{fig:objaverse}
\end{figure}

\yq{We provide some qualitative comparisons on Objaverse dataset in Figure~\ref{fig:objaverse}. }

\section{\yq{How UV unwrapping influence NBF}}
\yq{
In the proposed NBF unprojection strategy, a core step is to detect border areas. Since we conduct the border detection process in UV space, a natural question is: How does UV unwrapping influence this process? For example, what if a continuous region of a 3D mesh is partitioned into two disconnected charts in UV space? To answer these questions, we discuss from both experimental and theoretical perspectives:

\begin{itemize}
    \item Our experiments show that  Xatlas provides relatively robust UV unwrapping, enabling NBF to outperform baselines both qualitatively and quantitatively, as shown in the experimental results of our manuscript in Section 4.4 ``Ablation Study on NBF Unprojection''.
    
    \item Theoretically, when a single region of a 3D mesh is partitioned into two disconnected charts in UV space, NBF unprojection roughly degenerates to naive unprojection within the additional mis-segmented boundary region. Consequently, the lower performance bound of NBF is approximately equivalent to that of the naive unprojection. The detailed derivation is provided below.
    
\end{itemize}
When a single region of a 3D mesh is partitioned into two disconnected charts in UV space, this results in an additional segmentation boundary region that now corresponds to ``chart borders'' (dilated chart edges) in UV space. Crucially, for any view where these chart borders are visible, chart borders inherently belong to visibility borders, since they are near empty (invisible) areas that belong to no chart. Consequently, the border detection step does not distinguish these views' priority (now that they are all border areas), and subsequent steps rely solely on their directional priority. Therefore, NBF unprojection for such additional segmentation boundary regions becomes equivalent to the naive unprojection without considering border prioritization.
}

\clearpage

\begin{table}[t]
  \centering
  \caption{Geometry evaluation results of our adopted POCO and baseline methods. Note that baseline Texture Field also uses POCO for geometry extraction.}
  \vspace*{-4mm}
  \resizebox{\linewidth}{!}{
    \begin{NiceTabular}{c|c|ccc}
\toprule
 \multicolumn{1}{c|}{\textbf{ShapeNet Cat.}} & \textbf{Method} & \multicolumn{1}{c}{\textbf{CD ↓}} & \multicolumn{1}{c}{\textbf{NC ↑}} & \multicolumn{1}{c}{\textbf{FS ↑}} \\
    \midrule
    \multicolumn{1}{c|}{\multirow{3}[2]{*}{Chair}} & SPR   & 1.3094 & 0.9024 & 0.8710 \\
          & NKSR  & 0.9079 & 0.8694 & 0.8188 \\
          & POCO (TF \& Ours) & \cellcolor[rgb]{ .906,  .902,  .902} \textbf{0.4349} & \cellcolor[rgb]{ .906,  .902,  .902} \textbf{0.9267} & \cellcolor[rgb]{ .906,  .902,  .902} \textbf{0.9632} \\
    \midrule
    \multicolumn{1}{c|}{\multirow{3}[2]{*}{Car}} & SPR   & 0.8428 & \cellcolor[rgb]{ .906,  .902,  .902} \textbf{0.8847} & 0.8945 \\
          & NKSR  & 0.7876 & 0.8201 & 0.8128 \\
          & POCO (TF \& Ours) & \cellcolor[rgb]{ .906,  .902,  .902} \textbf{0.4407} & 0.8677 & \cellcolor[rgb]{ .906,  .902,  .902} \textbf{0.9462} \\
    \midrule
    \multicolumn{1}{c|}{\multirow{3}[2]{*}{Motirbike}} & SPR   & 4.2807 & 0.7647 & 0.6587 \\
          & NKSR  & 0.8786 & 0.6871 & 0.7249 \\
          & POCO (TF \& Ours) & \cellcolor[rgb]{ .906,  .902,  .902} \textbf{0.4423} & \cellcolor[rgb]{ .906,  .902,  .902} \textbf{0.7663} & \cellcolor[rgb]{ .906,  .902,  .902} \textbf{0.9434} \\
    \midrule
    \midrule
    \multicolumn{1}{c|}{\textbf{Dataset}} & \textbf{Method} & \multicolumn{1}{c}{\textbf{CD ↓}} & \multicolumn{1}{c}{\textbf{NC ↑}} & \multicolumn{1}{c}{\textbf{FS ↑}} \\
    \midrule
    \multicolumn{1}{c|}{\multirow{3}[2]{*}{GSO}} & SPR   & 0.8565 & 0.9446 & 0.9265 \\
          & NKSR  & 0.6149 & 0.9296 & 0.9006 \\
          & POCO (TF \& Ours) & \cellcolor[rgb]{ .906,  .902,  .902} \textbf{0.4424} & \cellcolor[rgb]{ .906,  .902,  .902} \textbf{0.9474} & \cellcolor[rgb]{ .906,  .902,  .902} \textbf{0.9679} \\
    \midrule
    \multicolumn{1}{c|}{\multirow{3}[2]{*}{OmniObject3D}} & SPR   & 0.6777 & 0.9511 & 0.9407 \\
          & NKSR  & 0.6148 & 0.9424 & 0.9003 \\
          & POCO (TF \& Ours) & \cellcolor[rgb]{ .906,  .902,  .902} \textbf{0.3393} & \cellcolor[rgb]{ .906,  .902,  .902} \textbf{0.9669} & \cellcolor[rgb]{ .906,  .902,  .902} \textbf{0.9866} \\
    \bottomrule
   
    \end{NiceTabular}%
    
    }
  \label{tab:geo_eval}%
\end{table}%

\begin{table}[h]
  \centering
  \caption{Quantitative results of replacing our geometry extraction module from POCO to SPR and Depth Inpainting.}
  \vspace{-1em}
   \resizebox{\linewidth}{!}{
    \begin{NiceTabular}{c|cccc}
    \toprule
    Geometry Module & PSNR ↑ & SSIM ↑ & LPIPS ↓ & FID ↓ \\
    \midrule
    SPR   & 19.4071 & 0.8752 & 0.1777 & 74.9854 \\
    Depth Inpainting & 26.2502  & 0.9466  & 0.0717  & 20.7198  \\
    Our Adopted POCO  & \cellcolor[rgb]{ .906,  .902,  .902} \textbf{26.2565 } & \cellcolor[rgb]{ .906,  .902,  .902} \textbf{0.9516 } & \cellcolor[rgb]{ .906,  .902,  .902} \textbf{0.0574 } & \cellcolor[rgb]{ .906,  .902,  .902} \textbf{4.9326 } \\
    \bottomrule
    \end{NiceTabular}%
    }
 
  \label{tab:geo}%
\end{table}%

\begin{table}[h]
  \centering
\caption{
  Quantitative results of replacing our inpainting module from DDNM to nearest interpolation, linear interpolation, and DiffPIR.
  }
  \vspace{-1em}
  \resizebox{\linewidth}{!}{
        
        \begin{NiceTabular}{c|cccc}
    \toprule
    Inpainting Module & PSNR ↑ & SSIM ↑ & LPIPS ↓ & FID ↓ \\
    \midrule
    Nearest Interpolation & 26.1175 & 0.9457 & 0.0618 & 11.0205 \\
    Linear Interpolation & 26.1739 & 0.9474 & 0.0612 & 9.4725 \\
    DiffPIR & 26.1582 & 0.9456 & 0.0652 & 9.4823 \\
    Our Adopted DDNM  & \cellcolor[rgb]{ .906,  .902,  .902} \textbf{26.2565 } & \cellcolor[rgb]{ .906,  .902,  .902} \textbf{0.9516 } & \cellcolor[rgb]{ .906,  .902,  .902} \textbf{0.0574 } & \cellcolor[rgb]{ .906,  .902,  .902} \textbf{4.9326 } \\
    \bottomrule
    \end{NiceTabular}%
    }
  
  \label{tab:inpaint}%
\end{table}%

\clearpage
\begin{table*}[t]
  \centering
  \caption{Quantitative results of using different $K$ values (numbers of viewpoints for projection and inpainting) on the motorbike category of ShapeNetCoreV2 dataset. More views contribute to a slightly higher reconstruction quality.
  }
    \begin{NiceTabular}{cc|cccc}
    \toprule
    Viewpoint Number & \multicolumn{1}{c|}{Camera Distribution} & PSNR ↑ & SSIM ↑ & LPIPS ↓ & FID ↓ \\
    \midrule
    6     & At the centers of each face of a cube & 20.8755 & 0.9273 & 0.0585 & 33.2870 \\
    8     & Evenly distributed on a Fibonacci Sphere & 21.0664 & 0.9287 & 0.0572 & 30.2113 \\
    20    & On the 20 vertices of a regular icosahedron & \cellcolor[rgb]{ .906,  .902,  .902} \textbf{21.2013 } & \cellcolor[rgb]{ .906,  .902,  .902} \textbf{0.9299 } & \cellcolor[rgb]{ .906,  .902,  .902} \textbf{0.0562 } & \cellcolor[rgb]{ .906,  .902,  .902} \textbf{28.4213 } \\
    \bottomrule
    \end{NiceTabular}%

  \label{tab:view_num}%
\end{table*}%

\end{document}